%% file: main.tex
\documentclass{article}

\usepackage[preprint]{corl_2025} %

\usepackage{times}
\input{preamble}

\definecolor{orange}{rgb}{1,0.5,0}
\definecolor{internationalorange}{rgb}{1.0, 0.31, 0.0}

\input{defs}

\begin{document}

\title{Wheeled Lab: Modern Sim2Real for Low-cost, Open-source Wheeled Robotics}

\author{
  Tyler Han$^{\ast}$ \quad \textbf{Preet Shah} \quad \textbf{Sidharth Rajagopal} \quad \textbf{Yanda Bao}\\[0.1cm]
  \textbf{Sanghun Jung} \;\; \textbf{Sidharth Talia} \;\; \textbf{Gabriel Guo} \;\; \textbf{Bryan Xu} \;\; \textbf{Bhaumik Mehta}\\[0.1cm]
  \textbf{Emma Romig} \;\; \textbf{Rosario Scalise} \;\; \textbf{Byron Boots}\;\;\\[0.3cm]
  \large University of Washington
}

\makeatletter
\let\@oldmaketitle\@maketitle%
\renewcommand{\@maketitle}{\@oldmaketitle%
    \centering
    \vspace{-0.7cm}
    \includegraphics[width=.75\linewidth]{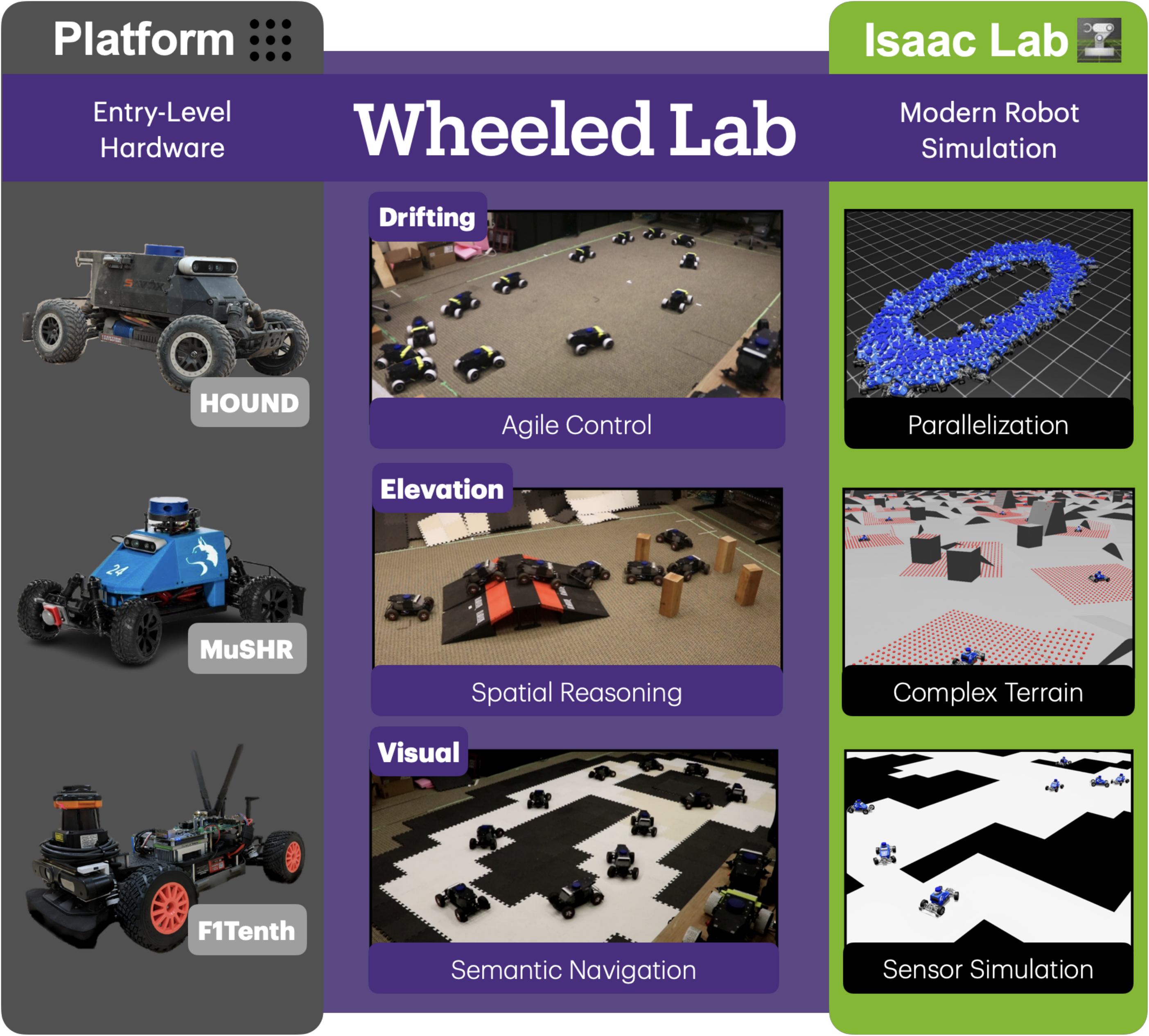}
    \captionof{figure}{Wheeled Lab bridges popular low-cost open-source wheeled platforms~\cite{talia_demonstrating_2024, srinivasa_mushr_2023, okelly_f1tenth_2020} with the research-backed robotics ecosystem Isaac Lab~\cite{mittal_orbit_2023} to support and democratize access to cutting-edge tools in modern robotics.}
    \label{fig:fig1}
    \vspace{-0.2cm}
}%
\makeatother

\maketitle

\begin{abstract}
Reinforcement Learning (RL) has been pivotal in recent robotics milestones and is poised to play a prominent role in the future.
However, these advances can rely on proprietary simulators, expensive hardware, and a daunting range of tools and skills. As a result, broader communities are disconnecting from the state-of-the-art; education curricula are poorly equipped to teach indispensable modern robotics skills involving hardware, deployment, and iterative development.
To address this gap between the broader and scientific communities, we contribute Wheeled Lab, an ecosystem which integrates accessible, open-source wheeled robots with Isaac Lab, an open-source robot learning and simulation framework, that is widely adopted in the state-of-the-art. To kickstart research and education, this work demonstrates three state-of-the-art zero-shot policies for small-scale RC cars developed through Wheeled Lab: controlled drifting, elevation traversal, and visual navigation. The full stack, from hardware to software, is low-cost and open-source. Videos and additional materials can be found at: \websiteliteral
\end{abstract}

\footnotetext[1]{Corresponding author: \texttt{than123@uw.edu}}

\section{Introduction}\label{sec:intro}
\vspace{-0.3cm}

\input{sections/intro} %

\section{Related Work}\label{sec:related}

\begin{table*}[t]
\renewcommand{\arraystretch}{1.5}
\footnotesize
\scalebox{.66}{
\begin{tabular}{lccccccccc}
\toprule
 & \multicolumn{3}{c}{\textbf{Sensor Simulation}} & \multicolumn{2}{c}{\textbf{Agent Physics}} & \multicolumn{4}{c}{\textbf{Reinforcement Learning}} \\
 \cmidrule(r){2-4} \cmidrule(r){5-6} \cmidrule(r){7-10}
 Ecosystem & \multicolumn{1}{c}{Elevation} & \multicolumn{1}{c}{LiDAR} & \multicolumn{1}{c}{Camera} & \multicolumn{1}{c}{Dynamics} & \multicolumn{1}{c}{Parallelization} & \multicolumn{1}{c}{Perturbation} & \multicolumn{1}{c}{Corruption} & \multicolumn{1}{c}{Community} & \multicolumn{1}{c}{Platforms} \\
 \midrule
Wheeled Lab & \cmark & 3D & Depth, RGB & \makecell{16 DOF \\+ Collisions} & High & \cmark & \cmark & RSL, SB3 & \makecell{HOUND, MuSHR,\\ F1Tenth} \\
AutoDrive~\cite{samak_autodrive_2023} & \xmark & 2D & \xmark & 16 DOF & Low & \xmark & \xmark & Gym & F1Tenth, Nigel \\
F1Tenth~\cite{okelly_f1tenth_2020}& \xmark & 2D & \xmark & Kinematic & \xmark & \xmark & \xmark & Gym & F1Tenth \\
CARLA~\cite{dosovitskiy_carla_2017}& \xmark & 3D & Depth, RGB & \makecell{16 DOF \\+ Collisions} & \xmark & \makecell{Steering\\Only} & \cmark & SB3 & \xmark \\
\shortstack[l]{Brunnbauer \textit{et al.}~\cite{brunnbauer_latent_2022}\\(PyBullet)} & \xmark & 2D & RGB & 12 DOF & \xmark & \xmark & \xmark & \xmark & F1Tenth \\
\shortstack[l]{Hamilton \textit{et al.}~\cite{hamilton_zero-shot_2022}\\(Gazebo)} & \xmark & 2D & \xmark & Kinematic &  \xmark & \xmark & \xmark & \xmark & F1Tenth \\
\shortstack[l]{\textit{PIETRA}~\cite{cai_pietra_2025}\\(Chrono)} & \cmark & \xmark & \xmark & 12 DOF & \xmark & \xmark & \xmark & \xmark & F1Tenth \\
RoboTHOR~\cite{deitke_robothor_2020} & \xmark & \xmark & RGB & Kinematic & \xmark & \xmark & \cmark & AllenAct & Household \\
AWS Deepracer~\cite{balaji_deepracer_2020} & \xmark & \cmark & RGB & Kinematic & \xmark & \xmark & \xmark & AWS & Deepracer \\
\bottomrule
\end{tabular}
}
\caption{\textbf{Comparisons of existing ecosystems on their capabilities.}  Various simulation, learning, and deployment ecosystems have been integrated with accessible wheeled platforms for Sim2Real. However, these ecosystems are noticeably isolated from the research community and, where implemented, have outdated features.
}
\label{tab:integrations}
\vspace{-15pt}
\end{table*}
\vspace{-10pt}
\input{sections/related} %

\begin{figure*}[t]
    \centering
    \includegraphics[width=.7\linewidth]{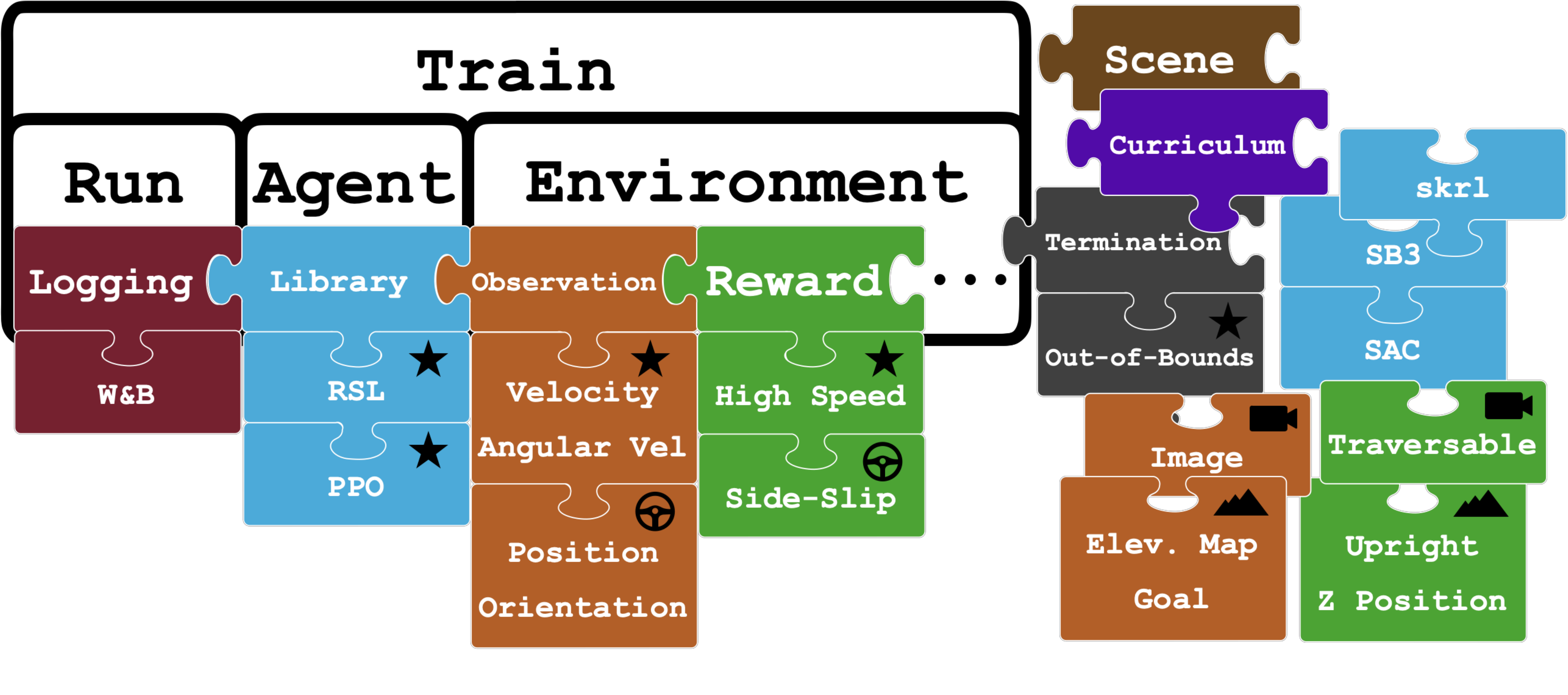}
    \vspace{-0.4cm}
    \caption{\textbf{Modular training framework imagined as the assembly of a puzzle.} Three main components comprise training: \texttt{Run}, \texttt{Agent}, \texttt{Environment}. Outlined puzzle pieces (e.g., \texttt{Observation}, \texttt{Reward}) are sub-components whose behaviors are defined by the pieces below them. For instance, \texttt{High Speed} is a reward setting in all of our RL environments, as denoted by the star (\includegraphics[height=\fontcharht\font`B]{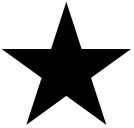}) icon.
    Wheel (\includegraphics[height=\fontcharht\font`B]{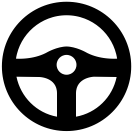}), camera (\includegraphics[height=\fontcharht\font`B]{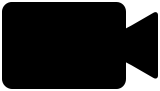}), and mountain (\includegraphics[height=\fontcharht\font`B]{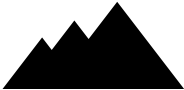}) icons indicate specific design choices for $\pdrift$, $\pvis$, and $\pelev$, respectively. Components shown are non-exhaustive.}
    \label{fig:overview}
\vspace{-10pt}
\end{figure*}

\section{Implementation}\label{sec:wheeled-lab}
\vspace{-6pt}
\input{sections/wheeled-lab} 
\vspace{-6pt}

\section{Experiments}\label{sec:experiments}
\vspace{-6pt}

\begin{wrapfigure}{l}{0.3\linewidth}
    \vspace{-12pt}
    \centering
    \includegraphics[width=\linewidth]{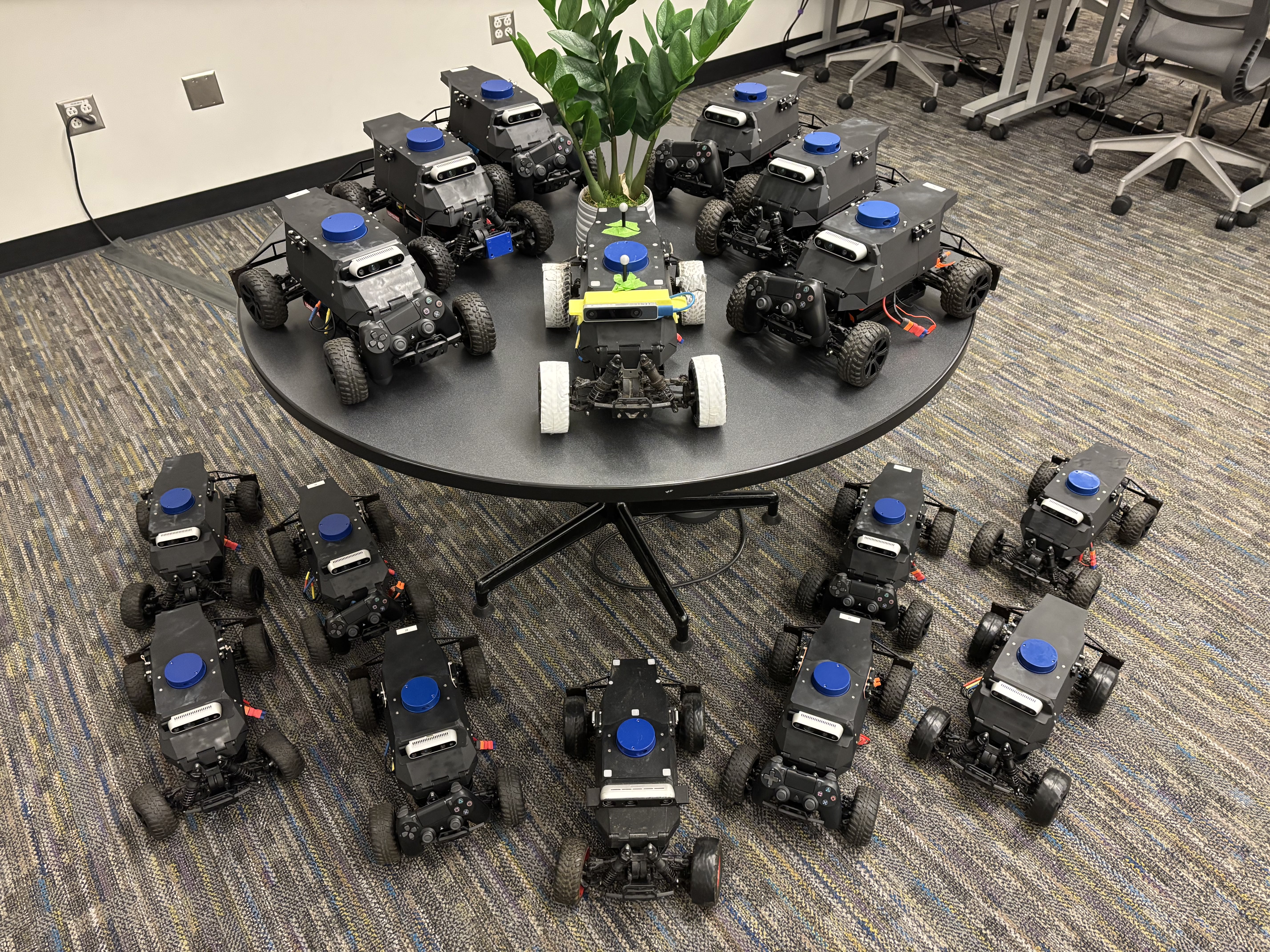}
    \caption{Several autonomy platforms shared among students in a robotics class.}
    \label{fig:hound-family}
    \vspace{-12pt}
\end{wrapfigure}
While we contribute the novelties in applied RL as discussed in \Cref{sec:intro}, our experiments aim to highlight the disparity in open-source methodologies between the broader and scientific communities. Thus, we compare policies trained through Wheeled Lab to baselines which reflect what training ecosystems currently provide to low-cost wheeled platforms, as shown in \Cref{tab:integrations}. This includes ablations over techniques such as domain randomization, perturbations, parallelization capacity, and architectures.
$\overline{(\cdot)}$ denotes the baseline version of a policy (e.g., $\bdrift$). 
Across all policies, we use Proximal Policy Optimization (PPO)~\cite{schulman_proximal_2017} as implemented in the RSL RL library to train the agent in simulation~\cite{lee_learning_2020}.

We implement, deploy, and evaluate three state-of-the-art policy types: $\pdrift$, $\pelev$, and $\pvis$ on the HOUND and MuSHR platforms.
Settings for drifting and elevation tasks are shown in \Cref{tab:drift-elev-ablation}.
All other configuration settings not mentioned (e.g., rewards, actuator parameters, articulations) can be assumed to be the same between $\boldsymbol{\pi}$ and $\overline{\boldsymbol{\pi}}$.

Our experiments employ the HOUND~\cite{talia_demonstrating_2024} for $\pdrift$ and the MuSHR~\cite{srinivasa_mushr_2023} for $\pelev$ and $\pvis$. The HOUND's estimated total price is about 3000 USD for a complete autonomy package with sensing (i.e., LiDAR, RGB) and compute (i.e., Jetson Orin NX). The MuSHR is estimated at 930 USD.
A single NVIDIA RTX 3080 GPU is used for training. Isaac Lab also offers cloud-based training for broader audiences without personal GPU access.

\begin{figure*}
    \centering
    \includegraphics[width=\linewidth]{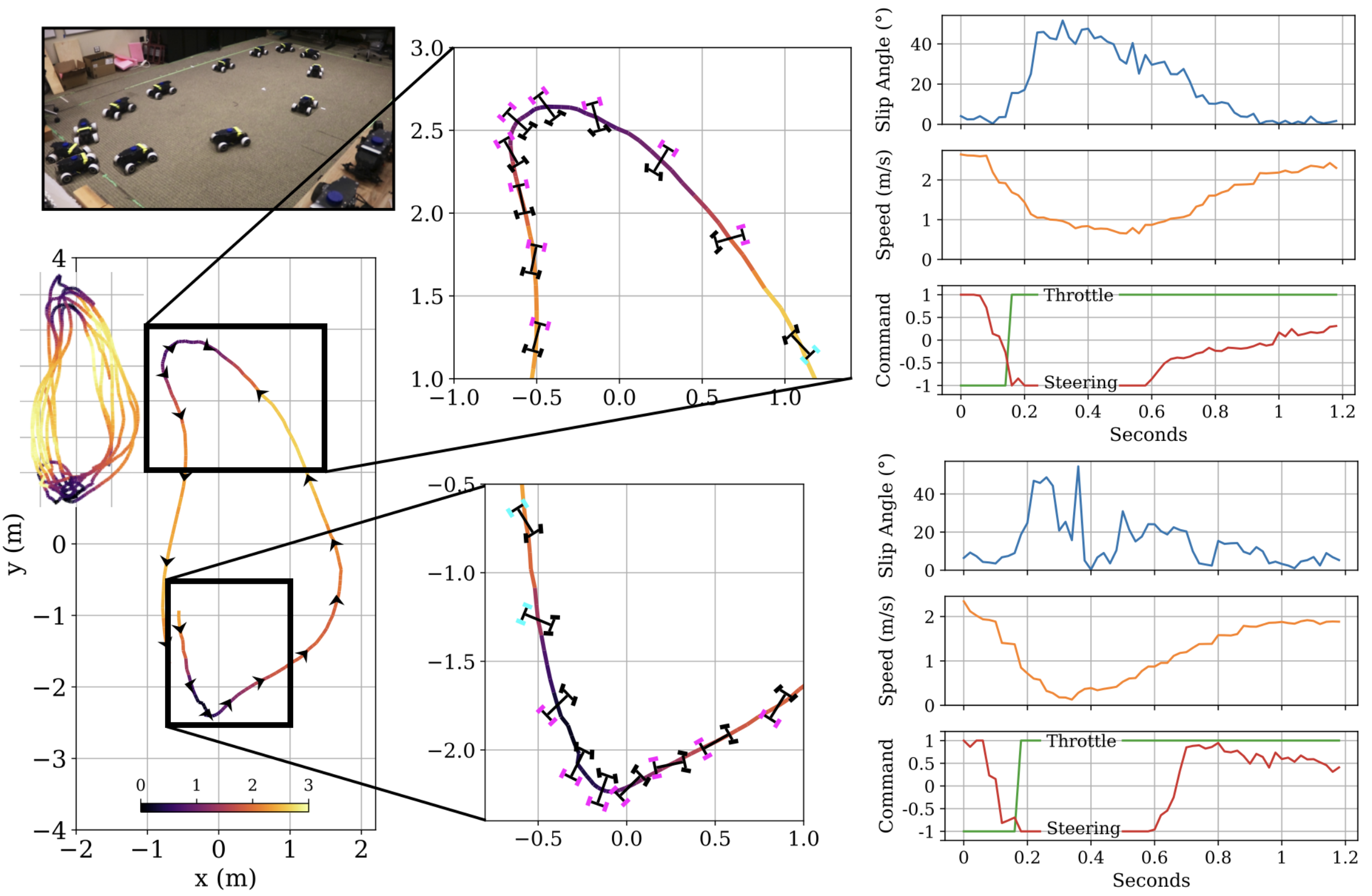}
    \vspace{-0.6cm}
    \caption{\textbf{A control strategy arises for drifting through turns.} \textbf{Left:} (Overlay: colored trajectories over multiple runs.) One run is spotlighted for this visualization. Arrows indicate the direction of travel, and the colorbar denotes speeds (in m/s). \textbf{Middle:} Magnification of drifted turns with markers visualizing vehicle orientation, steering (front wheel direction), and throttle (rear wheels; blue: $-1$, pink: $+1$). \textbf{Right:} Slip angle, speed, steering, and throttle are plotted for each turn. To initiate the drift, the platform quickly cuts the throttle to destabilize the rear wheels and then steers inwards sharply to throw them outwards. With the platform now facing the track center, it throttles through the remainder of the turn while counter-steering to control its residual angular momentum from the initial maneuver. The entire sequence occurs in just over 1 second. Visualization inspired by~\cite{djeumou_reference-free_2024}.}
    \label{fig:drift-vis}
    \vspace{-0.6cm}
\end{figure*}

\begin{wrapfigure}{r}{0.5\linewidth}
    \centering
    \includegraphics[width=\linewidth]{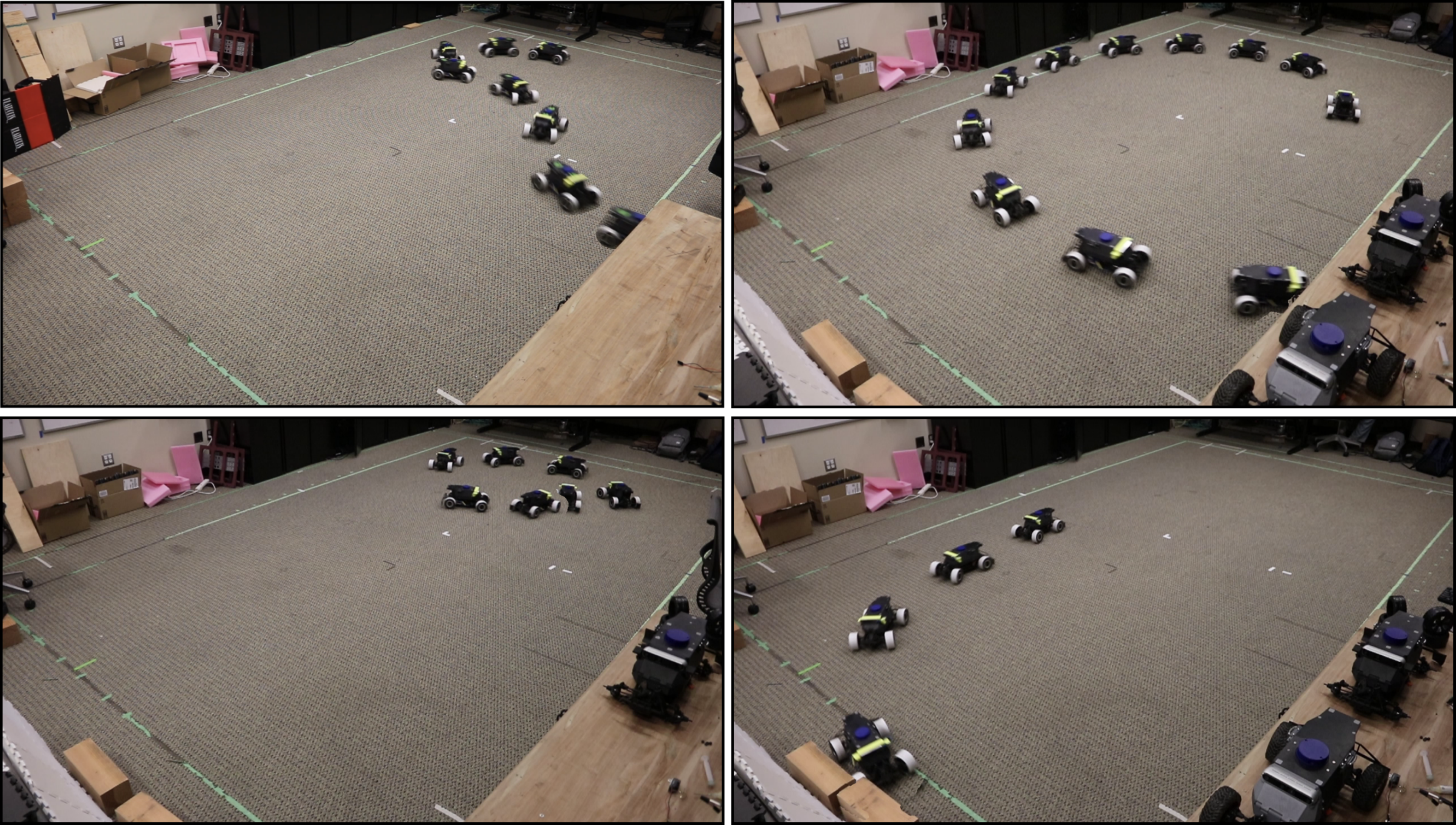}
    \vspace{-0.6cm}
    \caption{\textbf{Captured trajectories from the baseline drift policy in real experiments}. The baseline is unable to complete a valid turn without crashing or spinning out.}
    \label{fig:drift-baseline}
    \vspace{-0.6cm}
\end{wrapfigure}

\vspace{-0.3cm}
\subsection{Drifting}
\vspace{-0.3cm}
\textbf{Background.} Drifting is characterized by a vehicle's \textit{side-slip angle}, i.e., the angle between the heading and the velocity, as it maneuvers through a turn. In high-performance driving, drifting allows the vehicle to take much sharper turns at higher speeds than would be otherwise allowable by the no-slip turn radius~\cite{acosta_hybrid_2017}. The total mass, center-of-mass, friction properties of the surface and the wheels, actuator properties, state estimation, steering linkages, and suspension are all relevant factors in the narrow dynamics of a controlled drift.

The sensitivity of these dynamics becomes especially obvious when we (researchers) attempt the task manually despite the deceptively simple action space (see \website~for videos). When the maneuver is not executed correctly, the vehicle spins uncontrollably off track, losing traction (also known as a \textit{spin out}) and usually requiring a reset.

\textbf{Task.} For our evaluation, the task involves maintaining speed while minimizing the cross-track distance from an oval racing line without spinning out. 

\textbf{Setup.} State estimation is provided primarily by onboard Visual-Inertial-Odometry (VIO). We use a motion capture system for evaluation and occasional (1 hz) integrator-drift correction due to space limitations of our indoor testing environment. Successful VIO-only runs not used for data collection are shown on our \website. To enable the platform to be physically capable of drifting, tape is wrapped around the platform's rubber wheels to reduce friction and converted the base to a rear-wheel drive.

\begin{wraptable}{r}{.44\linewidth}
    \centering
    \vspace{-0.6cm}
    \scalebox{.66}{
    \begin{tabular}{l | c c c c c}
    \toprule
    Policy & Corruption & Perturbation & DR & \# Envs & Epochs \\
    \midrule
    $\bdrift$ & \cmark & \xmark & \xmark & 64 & 5000 \\
    $\pdrift$  & \cmark & \cmark & \cmark & 1024 & 5000 \\
    \midrule
    $\belev$ & \xmark & \xmark & \xmark & 128 & 1000 \\
    $\pelev$  & \cmark & \cmark & \cmark & 1024 & 1000 \\ 
    \bottomrule
    \end{tabular}
    }
    \vspace{-0.2cm}
    \caption{\textbf{Training settings}. Baseline policies $\pdrift$ and $\pelev$ (liberally) reflect capabilities currently available to low-cost, open-source wheeled platforms.}
    \vspace{-0.6cm}
    \label{tab:drift-elev-ablation}
\end{wraptable}
\textbf{Results.} We find that the baseline is unable to make a stable turn, much less complete a full lap (\Cref{fig:drift-baseline}). However, it occasionally attempts to counter-steer despite showing no indication during training. Baseline training runs consistently fail to discover the drifting mode. Motor cogging, noisy state estimates, constant slipping, and steering biases cause significant covariate shift.

On the other hand, $\pdrift$ is able to complete full laps. In addition, we observe evidence of high robustness in $\pdrift$ not yet seen in previous literature~\cite{williams_information_2017, djeumou_reference-free_2024}. That is, when the platform does spin out, instead of attempting to regain control by cutting the throttle, the policy maintains its angular momentum for a full spin (or more) before returning to the track. A visualization of the results (\Cref{fig:drift-vis}) gives insight into the policy's precise control. The maximum (controlled) slip angle is $58^\circ$, and the average speed is about $1.6$ m/s.
Appendix~\ref{app:drift-impl} provides further details about the design and implementation of $\pdrift$.

\vspace{-0.2cm}
\subsection{Elevation}
\vspace{-0.3cm}
\begin{figure}[htbp]
    \centering
    \begin{subfigure}[t]{0.4\linewidth}
        \centering
        \includegraphics[width=\linewidth]{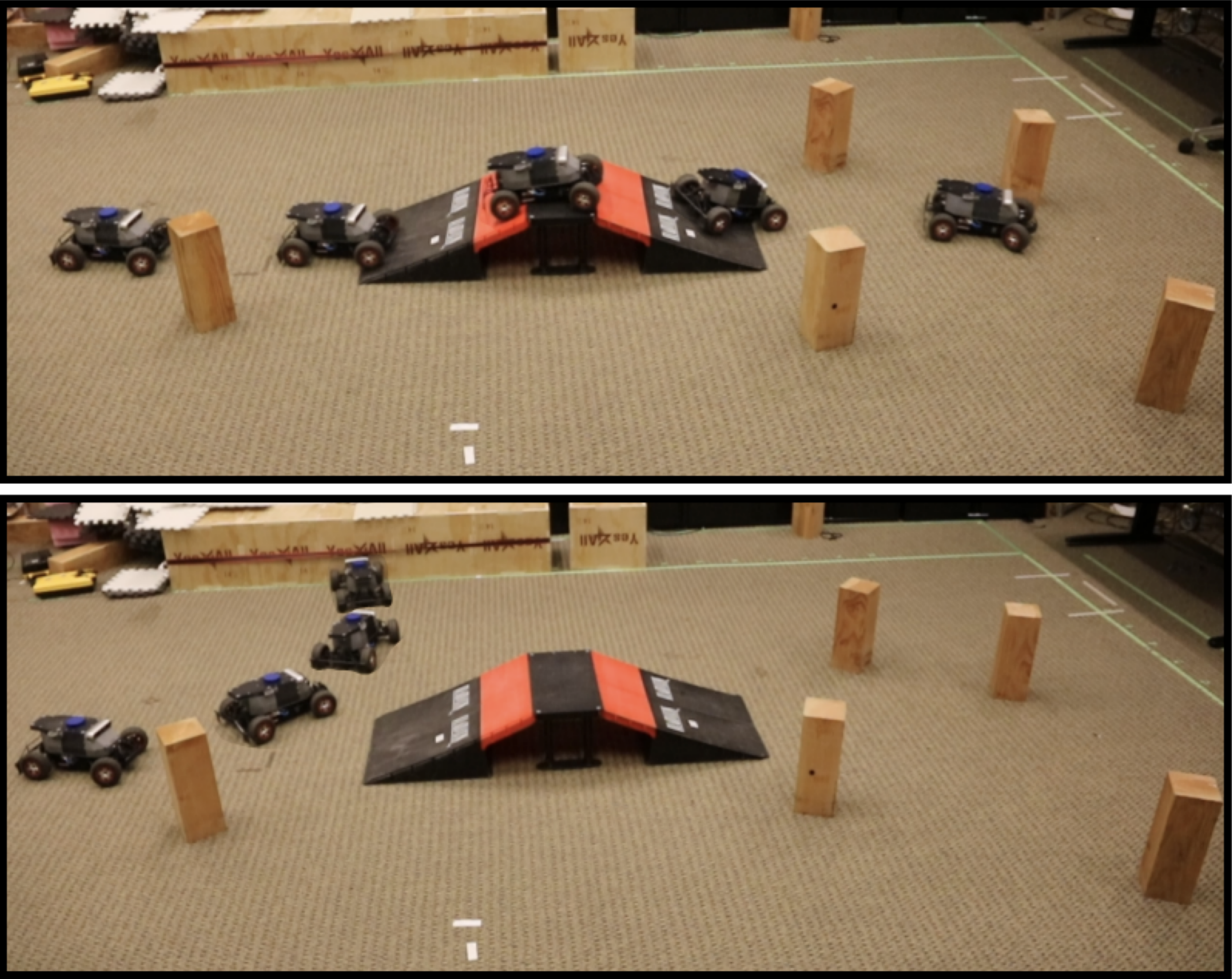}
        \caption{Captured trajectories of elevation policy in real experiments.}
        \label{fig:elev-real}
    \end{subfigure}
    \begin{subfigure}[t]{0.5\linewidth}
        \centering
        \includegraphics[width=\linewidth]{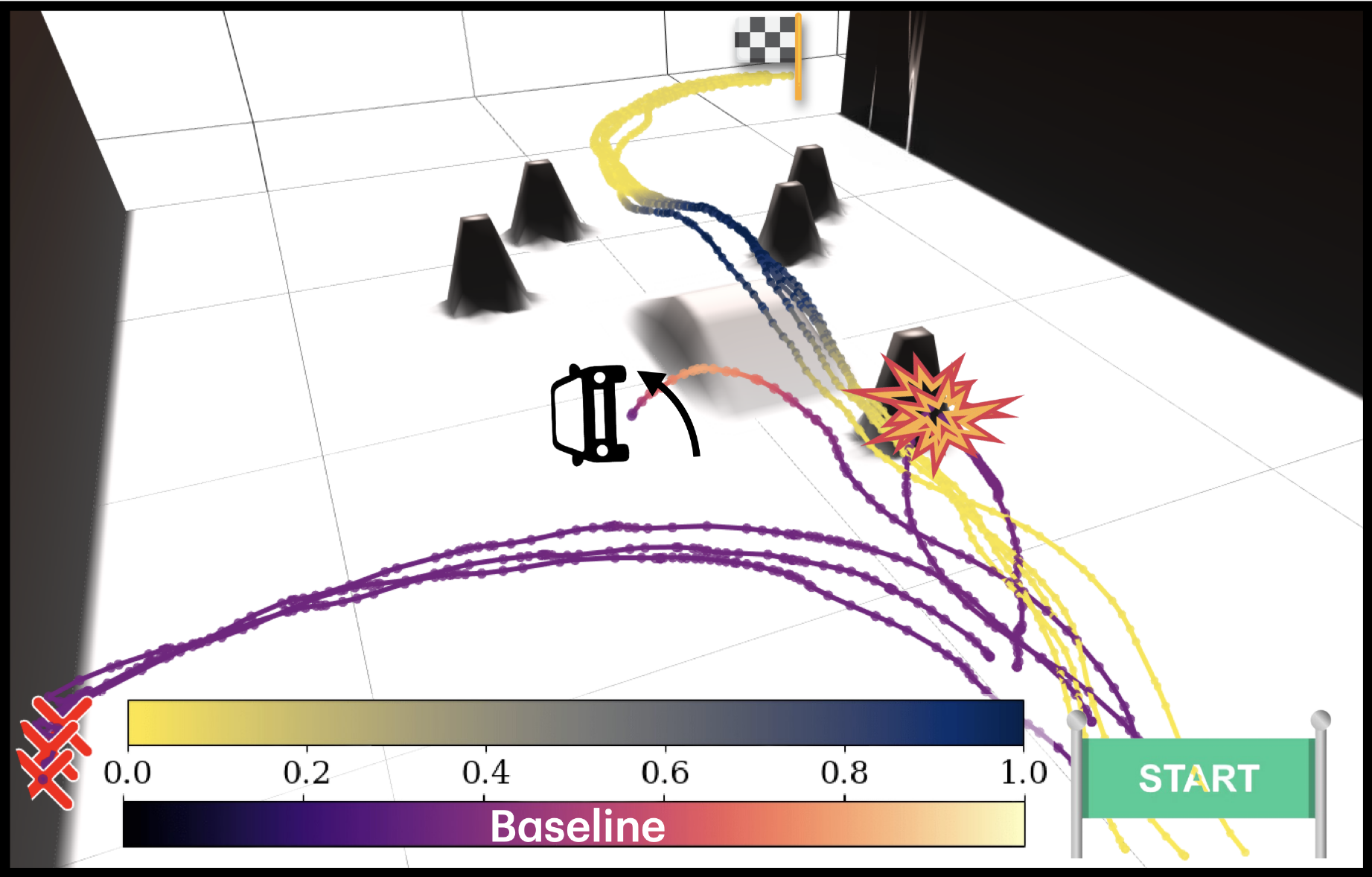}
        \caption{3D view of global elevation map.}
        \label{fig:elev-vis}
    \end{subfigure}
    \vspace{-0.1cm}
    \caption{
        Comparison of policy $\pelev$ and baseline $\belev$ in simulation and real-world elevation tasks.
        Fig.~\ref{fig:elev-real} shows real-world execution: $\pelev$(\textbf{top}) successfully avoids obstacles using local elevation, while $\belev$ (\textbf{bottom}) avoids elevation features entirely.
        Fig.~\ref{fig:elev-vis} visualizes the elevation map along with additional experiment result annotations. Trajectories from $\pelev$ (yellow/blue) and $\belev$ (purple/white) are overlaid. Policies only access their local $2.5\times2.5$ m maps during evaluation.
        }
    \vspace{-0.3cm}
    \label{fig:elev-subfigures}
\end{figure}

\textbf{Background.}
On unstructured and uneven terrain, an elevation map is essential for identifying traversable  areas in the scene. However, traversability is strongly dependent on agent morphology and dynamic state. Along with the ground geometry, the following affect how to safely traverse uneven terrain: the platform's ground clearance, center-of-mass height, wheel size, maximum torque, suspension, and momentum. Poor traversal can result in catastrophic failures, such as rollovers, and motor (or engine) stalling.

\textbf{Task.}
To show that the trained elevation-based policy is capable of geometric reasoning, we construct a scene with two types of elevation features that contribute to the overall elevation map, i.e., walls and ramps. Crucially, the incorporation of ramps sets the task apart from sole obstacle avoidance since ramps are features with height but are still traversable with the correct approach strategy.
The starting and goal positions are placed on opposite sides of the testing area, with the elevation features placed in between. The platform must safely traverse the scene to arrive at the goal position by avoiding obstacles and ascending ramps when locally feasible. 

\textbf{Setup.}
States provided to the policy are goal-relative position, orientation, and a local ($2.5\times2.5$ meters) body-centric elevation map. A local map is sampled from a prior global elevation map. Although the platform is capable of creating onboard maps~\cite{talia_demonstrating_2024}, a configurable indoor environment requires higher mapping resolution than is available for this evaluation.

\textbf{Results.} We find that the baseline primarily deviates from any elevation features to approach the goal. When the baseline does ascend the ramp, it quickly falls off the sides. However, $\pelev$ can both traverse the ramp safely and navigate through subsequent obstacles to reach the goal. Like $\bdrift$, the lower number of simulated agents gives the baseline policy significantly fewer opportunities to explore correct trajectories in a narrow range of safe states.

\vspace{-0.2cm}
\subsection{Visual}
\vspace{-0.2cm}
\begin{figure*}[!htb]
    \centering
    \includegraphics[width=0.8\linewidth]{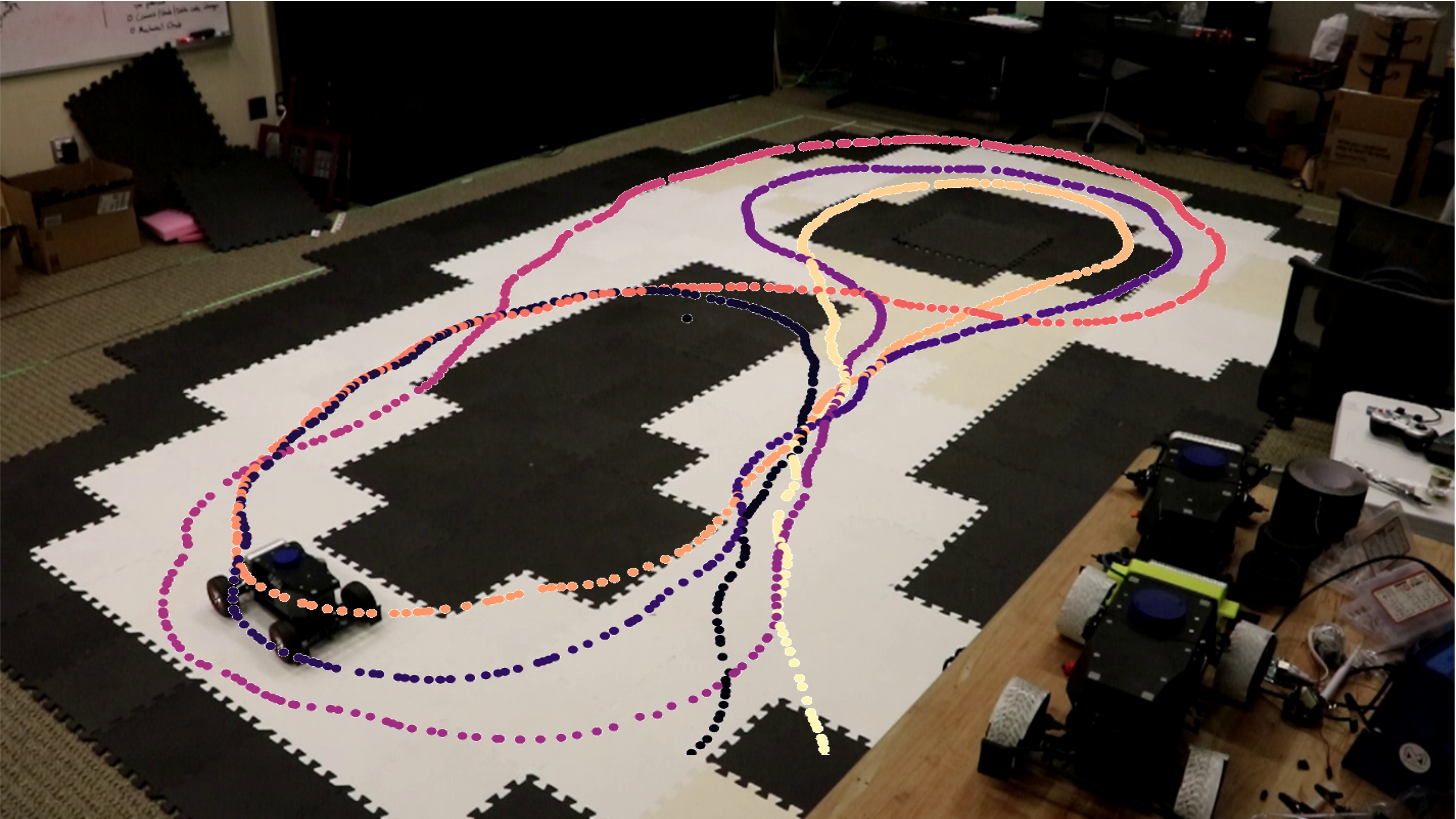}
    \vspace{-0.2cm}
    \caption{\textbf{Captured trajectories of visual policy in real experiments}. The trajectories are obtained from the motion capture system.
    We include videos of each setting and trial in the supplementary.}%
    \label{fig:visual_policy_trajectory}
    \vspace{-0.4cm}
\end{figure*}

\begin{wrapfigure}{r}{0.5\linewidth}
    \centering
    \vspace{-0.7cm}
    \includegraphics[width=\linewidth]{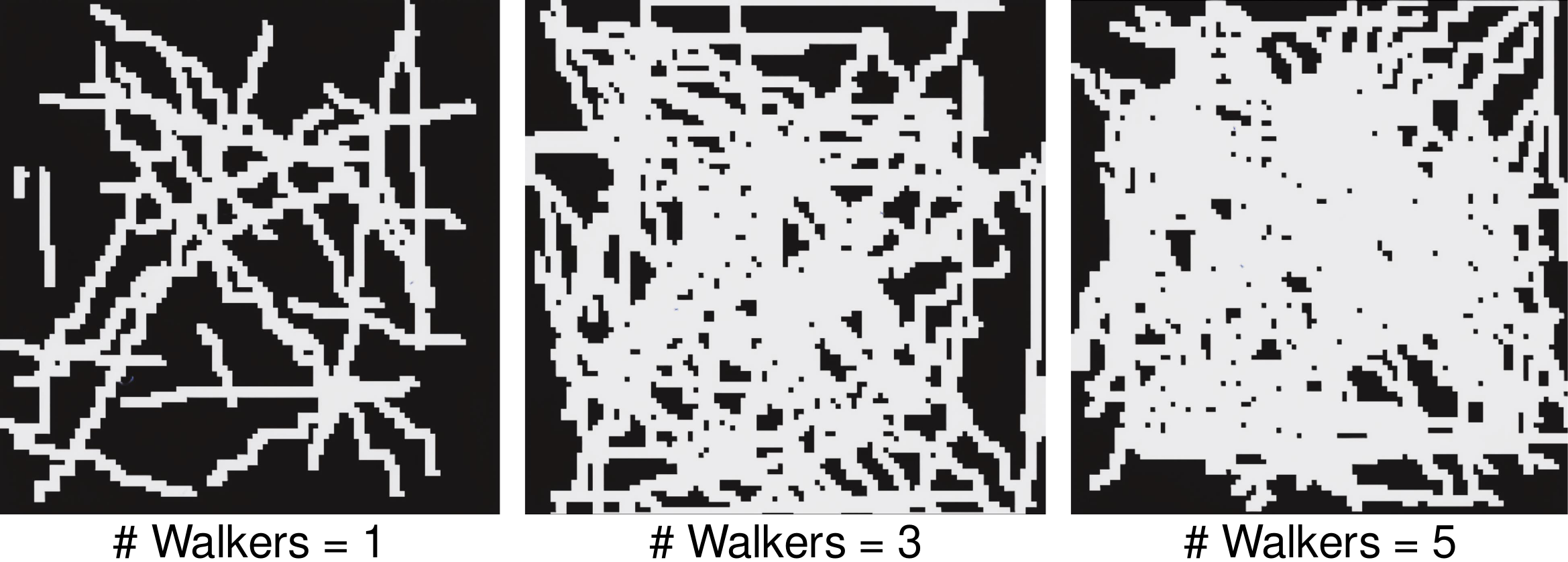}
    \vspace{-0.4cm}
    \caption{\textbf{Examples of randomly generated environments to train $\pvis$.} Black and white represent penalizing and safe regions, respectively. We adjust the number of walkers to train the policy across different difficulties. Using a smaller number of walkers will make the training environments more challenging.}
    \vspace{-0.3cm}
    \label{fig:visual_policy_maps}
\end{wrapfigure}
\textbf{Background.} Visual navigation traditionally relies heavily on a semantic understanding of the environment through visual feedback~\cite{jung_v-strong_2024}. Despite its importance, embedding information such as traversability through this modality has its unique challenges~\cite{yasuda_autonomous_2021}. 
First, visual observations (e.g., RGB images) from robots are represented in high-dimensional tensors, requiring exponentially more training data.
Also, small changes in environments that result from illumination, weather, or time of day can result in catastrophic failure of machine learning models~\cite{choi_robustnet_2021}.

These challenges can be effectively addressed through the simulation of visual information.
In addition to generating large amounts of task-relevant data, domain randomization can help improve generalization during transfer.

\textbf{Task.} We construct our task scene using black and white foam tiles (see Fig.~\ref{fig:visual_policy_trajectory}). These tiles are inexpensive, safe, and widely available. Surrounded by black tiles, a ``figure-8'' path is laid using white tiles. The platform must remain on the white path while avoiding the surrounding black tiles.
A dynamic version of this task is also used in evaluation; black ``barricade'' tiles are dynamically removed and placed to demonstrate robustness and real-time navigation reasoning.

We generate simulation environments with black and white areas, as we show in Fig.~\ref{fig:visual_policy_maps} to train the visual policy.
Each example in the figure represents a sub-environment, and our entire map is composed of sub-environments. This allows us to assign different difficulties for each sub-environment.
 See Appendix~\ref{app:vis} for additional details about map generation.

\textbf{Evaluation.} We measure success or failure over 
five trials.
We define success as the robot finishing at least one lap without staying in one place for more than 5 seconds.
\begin{wraptable}{r}{.5\linewidth}
    \centering
    \scalebox{.66}{
    \begin{tabular}{c|cc|cc}
        \toprule
        & \multicolumn{2}{c|}{MLP} & \multicolumn{2}{c}{CNN} \\
         & No Aug. & With Aug. & No Aug. & With Aug. \\
       \midrule
        \# of Success / Trial & 0 / 5 & 3 / 5 & 0 / 5 & 1 / 5 \\
        \bottomrule
    \end{tabular}
    }
    \caption{\textbf{Visual policy results.} Results demonstrate the effectiveness of image augmentations for Sim2Real generalization. While we expect CNN to be more capable of learning, MLP presents better generalization capability. We include videos of each run in the supplementary materials.}
    \vspace{-0.6cm}
    \label{tab:exp_visual_policy_results}
\end{wraptable}

\textbf{Setup.} Our input states include grayscale observation (resolution of $40\times60$), linear and angular velocity, and last action. Linear and angular velocities are obtained from the motion capture system.
We design experiments to demonstrate the effectiveness of image augmentations (i.e., color jittering and Gaussian blur) and different architectures (i.e., MLP vs CNN) in Sim2Real generalization.

\textbf{Results.} For quantitative evaluation, we run experiments five times and count the number of successes for each setting. As reported in Table~\ref{tab:exp_visual_policy_results}, image augmentation is essential in Sim2Real generalization. The policies without augmentation fail at generalization.
Further, despite the better learning capability of CNNs, they result in worse generalization to the real world; we also found that CNN training often collapses to a trivial solution, with the reward designs remaining the same as MLP training.
Such results imply that CNNs may overfit to simulated data and require more careful design in rewards and loss functions for stable training. 

In the supplementary materials, we qualitatively evaluate the results using recorded videos. Although both policies with augmentation predict meaningful actions, the MLP-based policy generates smoother and precise actions. In contrast, the CNN-based policy often generates noisy actions, as shown in fast steering changes and audible motor cogging.

\vspace{-0.3cm}
\section{Conclusion}\label{sec:conclusion}
\vspace{-0.3cm}

\input{sections/conclusions}

\clearpage\newpage
\section{Limitations}
\input{sections/limitations}

\clearpage\newpage
\bibliography{references}

\clearpage
\newpage
\input{sections/appendix}

\end{document}

%% file: preamble.tex
\usepackage[numbers]{natbib}
\usepackage{multicol}
\usepackage{rotating}

\usepackage[x11names]{xcolor}
\usepackage{amsmath}
\usepackage{graphicx}
\usepackage{cleveref}
\usepackage{dirtree}
\usepackage{caption}
\usepackage{amssymb}
\usepackage{pifont}
\usepackage{multirow}
\usepackage{booktabs}
\usepackage{scalerel}
\usepackage{float}
\usepackage{array}
\usepackage{makecell}
\usepackage{algorithm}
\usepackage{algpseudocode}
\usepackage{wrapfig}
\usepackage{subcaption}
\algnewcommand{\LeftComment}[1]{\Statex \(\triangleright\) #1}
\algnewcommand{\Do}[0]{\State \textbf{Do}}
\algnewcommand{\DoWhile}[1]{\State \textbf{do While} #1}

\usepackage[symbol]{footmisc}

%% file: defs.tex
\newcommand{\pdrift}{\boldsymbol{\pi}_{\text{drift}}}
\newcommand{\pelev}{\boldsymbol{\pi}_{\text{elev}}}
\newcommand{\pvis}{\boldsymbol{\pi}_{\text{vis}}}

\newcommand{\bdrift}{\overline{\boldsymbol{\pi}}_{\text{drift}}}
\newcommand{\belev}{\overline{\boldsymbol{\pi}}_{\text{elev}}}

\newcommand{\cmark}{\ding{51}}%
\newcommand{\xmark}{\ding{55}}%

\newcommand{\website}{\href{https://uwrobotlearning.github.io/WheeledLab/}{website}} %

\newcommand{\websiteliteral}{\href{https://uwrobotlearning.github.io/WheeledLab/}{\ttfamily https://uwrobotlearning.github.io/WheeledLab/}}

%% file: sections/intro.tex
The robotics community has made remarkable strides in recent years, boasting advances that enable quadrupeds to hike mountain trails and perform parkour~\cite{miki_learning_2022, hoeller_anymal_2024}, autonomous drones to race at champion-level ability~\cite{kaufmann_champion-level_2023}, and robot hands to solve Rubik's Cubes~\cite{openai_solving_2019}. Shared across these feats is simulated policy learning followed by direct deployment in the real world, termed \textit{Sim2Real}.

However, these remarkable Sim2Real innovations currently target expensive, space-intensive, and high-maintenance systems, including quadrupeds, humanoids, drones, and dexterous manipulators. Such platforms make it difficult for the general population of robotics builders and users to access the state-of-the-art techniques they employ, such as domain randomization and sensor simulation~\cite{kaufmann_champion-level_2023, huang_datt_2023, liao_berkeley_2024, hoeller_anymal_2024}.

In contrast, small-scale autonomous RC cars offer low-cost\footnote{About 3000 USD or less. See~\cite{samak_autodrive_2023} for a comparative analysis.} yet performative platforms that can leverage modern robotics methods, as we demonstrate in this paper. Used by classrooms, national racing competitions, and enthusiasts, 
platforms like F1Tenth~\cite{okelly_f1tenth_2020} have established themselves as go-to, entry-level platforms for beginning roboticists and researchers alike. Crucially, these platforms still provide sufficient onboard real estate to house components like the battery and compute necessary for modern algorithms, as well as sufficient mobility to exercise complex dynamics and environment interactions like drifting or off-roading~\cite{williams_information_2017, han_model_2024, han_dynamics_2024, datar_toward_2024}. In short, small-scale wheeled robots have high algorithmic potential at relatively low cost and hardware complexity.

A survey of existing ecosystems for low-cost wheeled platforms indicates that current simulation capabilities are isolated and outdated
(see \Cref{tab:integrations}). As a result, education~\cite{horvath_teaching_2024} and research opportunities~\cite{evans_unifying_2024} currently limit practitioners to model-based or model-free methods with low-fidelity simulation. With the advent of high-fidelity, open-source simulators like Isaac Lab~\cite{mittal_orbit_2023}, which are geared toward modern robotic learning, these limited ecosystems no longer need to be the status quo.

This work aims to support education and research by making modern robotics methods available to the broader community. To accomplish this, \textit{we contribute a Sim2Real framework for 
open-source wheeled robots that we have integrated with a state-of-the-art, research-grounded simulation ecosystem, Isaac Lab. }Modern methods we implement and evaluate include: massive parallelization, domain randomization, sensor simulation, and end-to-end learning. We demonstrate these methods using three policy types trained in simulation and deployed on low-cost, open-source platforms:

\begin{enumerate}
    \item The \textbf{Drifting Policy} ($\pdrift$) performs a controlled drift through extensive domain randomization, an approach in modern Sim2Real for tasks with uncertain and unstable dynamics. This is the first work to demonstrate direct Sim2Real transfer for drifting without online fine-tuning.
    \item The \textbf{Elevation Policy} ($\pelev$) traverses three-dimensional features with spatial reasoning. This demonstration highlights end-to-end training for modern tasks that tightly couple perception and control. Wheeled Lab is the first to provide integration of an accessible elevation-based Sim2Real pipeline with massive parallelization.
    \item The \textbf{Visual Policy} ($\pvis$) traverses visual features with semantic understanding. This demonstration highlights simulation, end-to-end training, and deployment with cameras, a sensor modality gaining substantial research attention due to its pre-training potential. Wheeled Lab provides a lower-cost alternative for visual Sim2Real compared to existing integrations.
\end{enumerate}

By integrating low-cost, open-source platforms with Isaac Lab, this work addresses a current need in robotics to extend modern principles in Sim2Real to resource-constrained audiences. 
\clearpage

%% file: sections/related.tex
Considerable prior work in Sim2Real has addressed open-source wheeled robots, including~\cite{datar_toward_2024, salimpour_sim--real_2025, balaji_deepracer_2020, samak_towards_2023, xiao_anycar_2024, cutler_autonomous_2016, williams_information_2017, evans_comparing_2023}. Some of these works produce one of $\pdrift$, $\pelev$, or $\pvis$ with partial or full Sim2Real. The individual accomplishments of this research both motivate our choice of demonstrations and help us highlight that \textit{varied state-of-the-art approaches can be coherently integrated under the framework presented here}.

\textbf{Drifting} \textbf{Policy}. Drifting is a challenging task in both optimal control and reinforcement learning (RL)~\cite{gonzales_autonomous_2016, cai_high-speed_2020, djeumou_reference-free_2024}. It is a fundamentally unstable maneuver whose control solutions are traditionally determined by vehicle and ground parameters~\cite{chen_dynamic_2023}. In many cases (ours included), researchers are unable to demonstrate a controlled drift for imitation-based methods~\cite{cutler_autonomous_2016}. 
These challenges make drifting an appealing task for RL. RL-based approaches have thus far been achievable only through further real-world fine-tuning~\cite{cutler_autonomous_2016, williams_information_2017}.
\textit{In this work, we show that aggressive domain randomization, perturbation simulation, and massive parallelization can enable direct transfer of a drifting policy.}
To reduce the barriers-to-entry, we deliberately distance our methods from techniques such as gain tuning and extensive system identification that require additional domain expertise and tooling.

\textbf{Elevation Policy.} It has become standard practice for locomotion policies on outdoor terrain to be trained with elevation maps~\cite{hoeller_anymal_2024}. 
In parallel, autonomous off-road vehicles have also demonstrated the importance of incorporating elevation in model-based methods~\cite{han_dynamics_2024, frey_roadrunner_2024, gibson_multi-step_2023}. Generally, these elevation-based tasks address challenges that arise at the interface of agent morphology and perception, which are often difficult to model. Therefore, 
lower-cost wheeled platforms have traditionally been oriented toward flat terrain~\cite{okelly_f1tenth_2020, balaji_deepracer_2020}, revealing a gap in methodology between the broader community and the state-of-the-art. Notable exceptions  
include~\cite{datar_toward_2024, stachowicz_fastrlap_2023, xu_reinforcement_2024} but do not offer as extensive community support for RL-based approaches.

\textbf{Visual Policy.} Improvements in camera and photorealistic scene simulation have advanced policies capable of semantic reasoning, bridging the gap between pre-trained visual models and embodied actions~\cite{kang_generalization_2019, yuan_learning_2024}.
Demonstrations of small-scale platforms with pre-trained visual models have also been investigated~\cite{stachowicz_fastrlap_2023}. In general, cameras provide an approachable and interpretable sensor modality for modern robotics education and research. Unfortunately, similar to elevation policies, visual policy pipelines on low-cost mobile platforms are not readily available even though the success of proprietary camera-based Sim2Real platforms suggests a clear demand~\cite{balaji_deepracer_2020}. \textit{This work demonstrates simulation, end-to-end training, and deployment with cameras on a low-cost platform.}

\textbf{Isaac Lab Simulation Framework.} This work builds on Isaac Lab, an open-source, widely adopted, and rapidly growing simulation framework for robotics research~\cite{mittal_orbit_2023, liao_berkeley_2024} that is heavily supported by industry and academia alike~\cite{technologies_1x_nodate, noauthor_crossing_nodate, noauthor_get_nodate, noauthor_nvidia_nodate, noauthor_menteebot_nodate, liao_berkeley_2024}. Isaac Lab has become a primary learning framework of choice for recent robotics research due to its impressive performance and ease-of-use features~\cite{ma_eureka_2024, hoeller_anymal_2024, tao_maniskill3_2024, yang_agile_2024, han_model_2025}.

%% file: sections/wheeled-lab.tex
Wheeled Lab bridges low-cost wheeled platforms with Isaac Lab (see \Cref{fig:fig1}). We extend Isaac Lab by designing infrastructure and assets which enforce reproducible experimentation. Some of the highest-level abstractions towards this effort are illustrated in \Cref{fig:overview} (e.g., Scene, Observation, Reward),
highlighting the inclusion of run parameters, architecture, and hyperparameters in addition to existing environment configurations. This design helps ensure reproducible, accessible, and portable experimentation settings and results, which is critical for practitioners to build upon existing work.

This work presents reliable and empirically validated baseline configurations of these abstractions, which are amenable specifically to low-cost wheeled platforms.
As Sim2Real remains an iterative process, especially in low-cost settings, our implementations consider accessibility factors throughout the entire development process: limited onboard compute and sensing, visual diversity, system identification feasibility, access to evaluation terrain, access to open-source packages, and access to training resources. Further implementation details for each task can be found in \Cref{app:implementation}.

%% file: sections/conclusions.tex
Drawing from our results in \Cref{sec:experiments},
observe that there is an alarming gap in techniques between the scientific and broader communities. Methods and modalities (e.g., parallelization, corruption, elevation), almost native in popular implementations and frameworks for platforms like humanoids and quadrupeds, are scattered throughout different implementations and research efforts for wheeled platforms. 

We present this work, Wheeled Lab, to act as a bridge between the broader community and state-of-the-art research in robot learning through low-cost, open-source wheeled platforms. Atop Isaac Lab, Wheeled Lab provides a modular and self-contained design framework to lower the barrier-to-entry while being fully integrated with Isaac Lab's extensive research community and ecosystem.

To kickstart this integration, we demonstrate the entire Sim2Real pipeline on three state-of-the-art policy types. Each policy and task is motivated by their fundamental relevance to general robot learning and Sim2Real. \textit{Drifting} is a complex control problem that highlights accurate, randomized, and massive physical simulation. Through Wheeled Lab, we demonstrate the first zero-shot transfer of a drifting policy thus far in the literature. \textit{Elevation traversal} defines safe behavior at the interface between perception and control, highlighting physical scene generation and massive interaction. \textit{Visual navigation} integrates a sensor modality with substantial research attention and pre-training potential while showcasing challenges in data diversity and transferability.

Wheeled Lab is entirely low-cost and open-source. Please refer to the project \website~(\websiteliteral) for experiment videos and more information.

%% file: sections/limitations.tex
Although an objective of Wheeled Lab is to support education, this work has not yet conducted a study to evaluate how comprehensible the pipeline is to robotics students. Like F1Tenth, we aim to construct an open curriculum around Wheeled Lab to help introduce difficult or interdisciplinary topics that may be sources of confusion. This study will help reveal what steps in the Sim2Real process might prove most challenging or unfamiliar.

While iterative testing for unique conditions can serve pedagogical purposes, the appeal of building, training, and deploying a platform that ``just works" can be inspirational for beginner roboticists. However, the policies presented here in their current open-source state must be met with patience and an intent to iterate. Thus, custom low-cost wheeled platforms invite yet another frontier of modern robot learning: robust adaptation~\cite{bousmalis_using_2018}. For instance, the development of a drifting pipeline that adequately adapts to friction on deployment can help lower the barrier-to-entry even further.

\subsection*{Towards Accessible Sim2Real}

The development of Wheeled Lab is constrained by both platform \textit{and development} to low-cost yet easily obtainable resources.
Mechanical hysteresis and tolerances of hobby-grade components, limited onboard compute and sensing, visual diversity, methodological simplicity, access to evaluation terrain, access to open-source packages, and access to training resources are all considered in the development of Wheeled Lab. Sim2Real accessibility is not only a challenging constraint on robot hardware but on the entire development process.

In accessibility efforts, experts are perhaps paradoxically limited by their expertise, making viewing problems from the perspective of a novice or non-traditional practitioner challenging or impossible.
We find that broader audiences tend to benefit significantly from approachable infrastructure, familiar abstractions, modularity, and straightforward, reproducible, yet instructive tasks and experiments. Wheeled Lab has fortunately garnered interest from the broader community in support of these objectives, receiving contributions and feedback from the perspective of non-scientific audiences. Meanwhile, we expect that experts are better positioned to contribute fundamental concepts with vision.
Wheeled Lab is an essential step towards aligning these communities and fostering a future of robot learning where all roboticists can equip themselves with indispensable skills in hardware, deployment, and iterative development.

%% file: sections/appendix.tex
\newcommand{\nt}[1]{\textcolor{Green4}{#1}} %
\newcommand{\uwt}[1]{\textcolor{DarkOrchid2}{#1}} %

\appendix

\section{Supplementary Material}

Videos and more information can be found on the project website:

{\small\websiteliteral}

\section{Design and Implementation Details}\label{app:implementation}

The typical ranges on friction, mechanical tolerances, damages, etc. for low-cost platforms are far from ideal. However, iterating and developing on individual platforms in unique environments is practically and pedagogically critical, especially for real-world robotics. We hope that this work allows practitioners to focus their efforts on precisely this development cycle rather than infrastructure. The following implementation details should be regarded as a starting point and potential reference for troubleshooting inevitable issues.

Isaac Lab environment configurations primarily include parameter settings and reward functions, and can be quickly parsed while efficiently communicating environment behavior. We highlight some notable details in this section.

\subsection{Drifting}\label{app:drift-impl}

We found that drifting was an exemplary task to demonstrate the physical Sim2Real gap. Much of the development cycle was focused on closing this gap, as supported by countless other seminal works~\cite{openai_solving_2019, hoeller_anymal_2024}. 

\subsubsection{Actuators} In many related works for agile tasks, data is typically collected to assess actuator response and estimate gains. However, this process can be unfamiliar to beginning roboticists or non-traditional enthusiasts. \textit{Instead, all actuator parameters are extracted exactly from their technical specification sheet online, and gains are domain-randomized across environments.} Achieving the task in this manner helps to convey that, while domain expertise can certainly improve performance (domain randomized policies are known to find conservative solutions~\cite{liao_berkeley_2024}), it is not strictly necessary through modern methods.

It is worth noting that a few ``Sim2Real2Sim'' (or Sim2Sim) cycles were spent narrowing the randomized gain ranges for the throttle actuator settings. Due to neural network policy tendencies to exhibit ``bang-bang'' (on-off) behavior, gain values can become extremely important to the actual forces exerted on the vehicle.

Establishing and improving pipelines for this process can be an accessible yet effective future contribution for use by potential practitioners.

\subsubsection{Friction}

Initially, the domain randomization over friction was used to cover a wide range (0.2 to 0.8) of potential friction values between the tape-covered tires and the carpeted ground. However, resulting policies would perform poorly and inconsistently on deployment. A long thread of Sim2Sim cycles was used to search for friction parameters but still resulting policies were not reliably transferring.

In the end, a cheap (less than 5 USD) spring scale was used to estimate the coefficient of friction, measured by dragging the vehicle with the scale along the carpet. This value (about 0.4) was then used as the midpoint of the friction randomization during training.

The project \website~also contains videos of deployments on different surfaces (e.g. finished concrete). Lower friction values appear to cause the vehicle to take wider turns (due to lack of turning friction) and spin out quickly during drift maneuvers. Enabling the platform to drift over multiple surfaces is also of fundamental interest to the robotics community as it is a form of domain adaptation.

\subsubsection{Articulations}

While we initially used the MuSHR~\cite{srinivasa_mushr_2023} articulations defined by the open-source URDF file, we found that its modeling inaccuracies affected the precise control expectations of the platform.
As a result, we improved the articulations: (1) by moving the steering linkages from the wheel axis to the actual steering joint and (2) by articulating the suspension joints of the vehicle.

Steering inaccuracies were noticeable due to the gap between turning radii in real and sim. The moment arm on the linkage can also adversely affect the tracking of the steering joint during high-velocity maneuvers. This can even be observed in simulation when steering joints behave strangely asymmetrically with unsuitable actuator settings.

Suspension helps close the gap on load shifting during turns. With a suspension travel of 4 cm, these shifts have consequences in the total friction and ``thrust'' produced while throttling through turns. Interestingly, suspension articulation also helped to stabilize the simulation. At a wheel speed of 3 m/s, the rotation rate of the wheels reaches 60 rad/s, occasionally resulting in collision penetrations during simulation which cause unrealistic ``jumping'' through drifted turns. Damped suspension joints helped to stabilize these collisions without requiring smaller simulation time steps which would significantly lengthen training times.

\subsubsection{Rewards}

Six rewards were designed for this task:

\begin{enumerate}
    \item[i.)] \textbf{Cross-Track Distance}. The lateral distance of the vehicle from an oval track line. This is defined without time-dependent reference trajectories as often done in model-based approaches or even in deep RL methods~\cite{cai_high-speed_2020}. This is highly weighted otherwise, the agent tends to take larger, safer turns.
    \item[ii.)] \textbf{Velocity}. Reward for high speeds.
    \item[iii.)] \textbf{Side-Slip}. Only stable side-slip angles were rewarded. Stable side-slip angles are manageable through the steering limits (0-30$^\circ$).
    \item[iv.)] \textbf{Progress}. Reward for making progress along an arbitrary direction (counter-clockwise, in this work).
    \item[v.)] \textbf{Turn-Energy}. Shaping term for increasing velocity rewards, specifically when inside turn regions.
    \item[vi.)] \textbf{Turn-Left-Go-Right}. Shaping term for encouraging exploration to discover an alternate turning mode (e.g., drifting) by counter-steering. Strictly positive when the angular velocity and the steering command are opposite to each other. 0 otherwise.
\end{enumerate}

Penalties were also given for going out-of-bounds.

\subsection{Elevation}

\subsubsection{Training}

Training elevation in simulation proved to be a challenging task. Training would often collapse into one of: (1) seeking goals while avoiding obstacles or (2) climbing ramps. Striking a balance between the two often required re-designing the scene or task configuration in simulation.

As observed for model-based methods~\cite{han_dynamics_2024}, evaluating uneven terrain traversal for vehicles can be challenging due to the reasonable alternate mode of avoiding risky terrain altogether, which may often be faster. However, the primary goal of this work is to better enable the broader community to engage with state-of-the-art methods in elevation-based robotics. Thus, the reward and scene were constructed to incentivize the agent to actively seek out risky terrain while penalizing failures.

Supported by machine learning literature, noise injection becomes less effective for generalization at higher dimensions. For states such as elevation maps, perceived noise is likely more structured than the additive Gaussians injected through observation corruption. While debugging, we found that the policy would improve if the elevation features were brought closer in shape to the features seen in the simulation. To resolve this without ``training on test'', the scene was augmented with slopes of various gradations and heights.

\subsubsection{Suspension}

As mentioned in Appendix \ref{app:drift-impl}, suspension joints were added to the open-source MuSHR articulations. This is critical for elevation where maintaining momentum up a ramp helps training stability. Without suspension articulation, climbing ramps have an abrupt and destabilizing impact at higher speeds.

\subsubsection{Rewards}
\begin{enumerate}
    \item[i.)] \textbf{Goal Velocity}. Rewards velocity projected onto the goal heading vector.
    \item[ii.)] \textbf{Z-Position}. Rewards larger z-positions (gained through ascending ramps).
    \item[iii.)] \textbf{Falling Penalty}. Penalizes negative body-frame z velocities.
    \item[iv.)] \textbf{Roll on elevation}. Penalizes roll angles while on elevation features.
\end{enumerate}

\subsection{Visual}\label{app:vis}
While visual information is essential in modern robotics to solve complex real-world problems, simulating such information requires state-of-the-art rendering techniques to close the Sim2Real gap.
Despite such computationally heavy efforts, machine learning models trained with simulated data often fail at real-world generalization due to diverse types of environmental changes, e.g., illuminations, weather, etc.
Instead, in this work, we ``simplify'' visual information from both simulation and real by compressing RGB pixels to grayscale, closing the Sim2Real gap.
Still, we observed that such compressed visual information exhibits a non-trivial amount of Sim2Real gap.
This section describes how we tackle this problem by providing details of our environment, training augmentations, architectures, and reward designs.

\subsubsection{Visual map generation}
We designed the problem by encoding traversability into color information.
Black tiles are non-traversable regions while white indicates safe to travel.
We divided the whole map into multiple sub-environments that are individually configurable. 
Then, we randomized white paths over the black area in each sub-environment.
Specifically, each sub-environment comprises 9 cells, and we sample a start point in each cell and generate paths toward random endpoints in the sub-environment.
We made the number of random paths from a single start point configurable, allowing the users to adjust the difficulty of the training (see Fig.~\ref{fig:visual_policy_maps}).
This can later be used for curriculum learning by assigning different difficulties for sub-environments.
We tested the generated map with more than 2,000 agents in parallel, running seamlessly with a single Nvidia RTX 3080 GPU.
The algorithm overview for sub-environment generation is presented in Alg.~\ref{alg:visual_map_generation}.
\begin{algorithm}[!ht]
\centering
\footnotesize
\caption{Visual Policy Sub-environment Generation} 
\begin{algorithmic}[1]
\State \textbf{Input:}
\State $(H_\text{env}, W_\text{env}) \gets$ \# of rows and \# of columns for each sub environment
\State $(H_\text{cell}, W_\text{cell}) \gets$ \# of rows and \# of columns for each cell
\State $N_\text{walkers} \gets$ \# of walkers for path generation 
\State
\State \textbf{Pseudo-code:}
\LeftComment{Initialize traversability map; 0 - not traversable / 1 - traversable}
\State $\mathbf{T}\in\{0, 1\}^{H_\text{env}\times W_\text{env}} \gets \mathrm{\mathbf{0}}$ 
\State \LeftComment{Sample start points from each cell}
\State $\mathbf{P}^\text{start} \gets [\,]$ \Comment{Initialize an array to save start points}
\LeftComment{Sample random start points for each cell}
\For{row $r=0, 1, \ldots, H_\text{env} / H_\text{cell}$}
  \For{col $c=0, 1, \ldots, W_\text{env} / W_\text{cell}$}
    \State $p_\text{row} \gets \text{UniformSampling}(0, H_\text{cell}) + r \cdot H_\text{cell}$
    \State $p_\text{col} \gets \text{UniformSampling}(0, W_\text{cell}) + c \cdot W_\text{cell}$
    \State $\mathbf{P}^\text{start}\, \mathrel{+}= [(p_\text{row}, p_\text{col})]$ \Comment{Append the start point to $\mathbf{P}^\text{start}$}
  \EndFor
\EndFor
\Statex

\For{$(p_\text{row}, p_\text{col})$ in $\mathbf{P}^\text{start}$}
  \For{iter$=0, 1, \ldots, N_\text{walkers}$}
    \Statex \ \ \ \ \ \ \ \ \ \(\triangleright\) Sample random end points
    \Do
      \State \ \ \ $q_\text{row} \gets \text{UniformSampling}(0, H_\text{env})$
      \State \ \ \ $q_\text{col} \gets \text{UniformSampling}(0, W_\text{env})$
    \DoWhile{$\mathbf{T}(q_\text{row}, q_\text{col}) == 1$}
    \Statex \ \ \ \ \ \ \ \ \ \(\triangleright\) Generate random paths and update traversability map
    \State GenerateRandomPaths$(\mathbf{T},\, (p_\text{row}, p_\text{col}),\, (q_\text{row}, q_\text{col}))$
  \EndFor
\EndFor
\State \textbf{Return:} $\mathbf{T}$
\end{algorithmic} 
\label{alg:visual_map_generation}
\end{algorithm}

\subsubsection{Model Architectures}
We adopt a multi-layer perceptron (MLP) as our basic policy network.
For ablations, we add a simple 3-layer convolutional neural network (CNN) that serves as an image feature extractor before the policy networks.
The extracted features are concatenated to the rest of the observation terms, such as linear/angular velocities and last action, and fed into the MLP policy networks.
We observe that inducing CNN takes longer training time to learn optimal actions and often collapses into a trivial solution - agents decide to spin in a circle, only maximizing the velocity rewards.

\subsubsection{Image Augmentations}
To close the sim-to-real gap, we apply image augmentations for policy training. This includes aggressive color jittering (\textit{i.e.,} brightness, contrast, saturation, and hue adjustments) and Gaussian blur.
As demonstrated in both qualitative and quantitative evaluations, image augmentations play a crucial role in closing sim-to-real gaps.
Further augmentations or randomization can be easily applied to our code design.

\subsubsection{Rewards}
Two rewards were designed for this task
\begin{enumerate}
    \item[i.)] \textbf{Velocity Reward}. Rewards higher velocities
    \item[ii.)] \textbf{Traversability Reward and Penalty}. Rewards being on white regions, and penalizes it for being on black regions
\end{enumerate}

\subsubsection{Terminal Conditions}
We experimented with different terminations. In addition to the standard time-out and out-of-bounds terminal conditions, we added a terminal condition when the robot is on black. However, this often resulted in a lack of exploration, and policies that produced ``jerky'' or undesirable actions, such as idling at the initial location.

%% file: main.bbl
\begin{thebibliography}{54}
\providecommand{\natexlab}[1]{#1}
\providecommand{\url}[1]{\texttt{#1}}
\expandafter\ifx\csname urlstyle\endcsname\relax
  \providecommand{\doi}[1]{doi: #1}\else
  \providecommand{\doi}{doi: \begingroup \urlstyle{rm}\Url}\fi

\bibitem[Talia et~al.(2024)Talia, Schmittle, Lambert, Spitzer, Mavrogiannis, and Srinivasa]{talia_demonstrating_2024}
S.~Talia, M.~Schmittle, A.~Lambert, A.~Spitzer, C.~Mavrogiannis, and S.~S. Srinivasa.
\newblock Demonstrating {HOUND}: {A} {Low}-cost {Research} {Platform} for {High}-speed {Off}-road {Underactuated} {Nonholonomic} {Driving}, July 2024.
\newblock URL \url{http://arxiv.org/abs/2311.11199}.
\newblock arXiv:2311.11199 [cs].

\bibitem[Srinivasa et~al.(2023)Srinivasa, Lancaster, Michalove, Schmittle, Summers, Rockett, Scalise, Smith, Choudhury, Mavrogiannis, and Sadeghi]{srinivasa_mushr_2023}
S.~S. Srinivasa, P.~Lancaster, J.~Michalove, M.~Schmittle, C.~Summers, M.~Rockett, R.~Scalise, J.~R. Smith, S.~Choudhury, C.~Mavrogiannis, and F.~Sadeghi.
\newblock {MuSHR}: {A} {Low}-{Cost}, {Open}-{Source} {Robotic} {Racecar} for {Education} and {Research}, Dec. 2023.
\newblock URL \url{http://arxiv.org/abs/1908.08031}.
\newblock arXiv:1908.08031 [cs].

\bibitem[O’Kelly et~al.(2020)O’Kelly, Zheng, Karthik, and Mangharam]{okelly_f1tenth_2020}
M.~O’Kelly, H.~Zheng, D.~Karthik, and R.~Mangharam.
\newblock {F1TENTH}: {An} {Open}-source {Evaluation} {Environment} for {Continuous} {Control} and {Reinforcement} {Learning}.
\newblock In \emph{Proceedings of the {NeurIPS} 2019 {Competition} and {Demonstration} {Track}}, pages 77--89. PMLR, Aug. 2020.
\newblock URL \url{https://proceedings.mlr.press/v123/o-kelly20a.html}.
\newblock ISSN: 2640-3498.

\bibitem[Mittal et~al.(2023)Mittal, Yu, Yu, Liu, Rudin, Hoeller, Yuan, Singh, Guo, Mazhar, Mandlekar, Babich, State, Hutter, and Garg]{mittal_orbit_2023}
M.~Mittal, C.~Yu, Q.~Yu, J.~Liu, N.~Rudin, D.~Hoeller, J.~L. Yuan, R.~Singh, Y.~Guo, H.~Mazhar, A.~Mandlekar, B.~Babich, G.~State, M.~Hutter, and A.~Garg.
\newblock Orbit: {A} {Unified} {Simulation} {Framework} for {Interactive} {Robot} {Learning} {Environments}.
\newblock \emph{IEEE Robotics and Automation Letters}, 8\penalty0 (6):\penalty0 3740--3747, June 2023.
\newblock ISSN 2377-3766, 2377-3774.
\newblock \doi{10.1109/LRA.2023.3270034}.
\newblock URL \url{http://arxiv.org/abs/2301.04195}.
\newblock arXiv:2301.04195 [cs].

\bibitem[Miki et~al.(2022)Miki, Lee, Hwangbo, Wellhausen, Koltun, and Hutter]{miki_learning_2022}
T.~Miki, J.~Lee, J.~Hwangbo, L.~Wellhausen, V.~Koltun, and M.~Hutter.
\newblock Learning robust perceptive locomotion for quadrupedal robots in the wild.
\newblock \emph{Science Robotics}, 7\penalty0 (62):\penalty0 eabk2822, Jan. 2022.
\newblock \doi{10.1126/scirobotics.abk2822}.
\newblock URL \url{https://www.science.org/doi/abs/10.1126/scirobotics.abk2822}.
\newblock Publisher: American Association for the Advancement of Science.

\bibitem[Hoeller et~al.(2024)Hoeller, Rudin, Sako, and Hutter]{hoeller_anymal_2024}
D.~Hoeller, N.~Rudin, D.~Sako, and M.~Hutter.
\newblock {ANYmal} parkour: {Learning} agile navigation for quadrupedal robots.
\newblock \emph{Science Robotics}, 9\penalty0 (88):\penalty0 eadi7566, Mar. 2024.
\newblock \doi{10.1126/scirobotics.adi7566}.
\newblock URL \url{https://www.science.org/doi/full/10.1126/scirobotics.adi7566}.
\newblock Publisher: American Association for the Advancement of Science.

\bibitem[Kaufmann et~al.(2023)Kaufmann, Bauersfeld, Loquercio, Müller, Koltun, and Scaramuzza]{kaufmann_champion-level_2023}
E.~Kaufmann, L.~Bauersfeld, A.~Loquercio, M.~Müller, V.~Koltun, and D.~Scaramuzza.
\newblock Champion-level drone racing using deep reinforcement learning.
\newblock \emph{Nature}, 620\penalty0 (7976):\penalty0 982--987, Aug. 2023.
\newblock ISSN 1476-4687.
\newblock \doi{10.1038/s41586-023-06419-4}.
\newblock URL \url{https://www.nature.com/articles/s41586-023-06419-4}.
\newblock Publisher: Nature Publishing Group.

\bibitem[OpenAI et~al.(2019)OpenAI, Akkaya, Andrychowicz, Chociej, Litwin, McGrew, Petron, Paino, Plappert, Powell, Ribas, Schneider, Tezak, Tworek, Welinder, Weng, Yuan, Zaremba, and Zhang]{openai_solving_2019}
OpenAI, I.~Akkaya, M.~Andrychowicz, M.~Chociej, M.~Litwin, B.~McGrew, A.~Petron, A.~Paino, M.~Plappert, G.~Powell, R.~Ribas, J.~Schneider, N.~Tezak, J.~Tworek, P.~Welinder, L.~Weng, Q.~Yuan, W.~Zaremba, and L.~Zhang.
\newblock Solving {Rubik}'s {Cube} with a {Robot} {Hand}, Oct. 2019.
\newblock URL \url{http://arxiv.org/abs/1910.07113}.
\newblock arXiv:1910.07113 [cs].

\bibitem[Huang et~al.(2023)Huang, Rana, Spitzer, Shi, and Boots]{huang_datt_2023}
K.~Huang, R.~Rana, A.~Spitzer, G.~Shi, and B.~Boots.
\newblock {DATT}: {Deep} {Adaptive} {Trajectory} {Tracking} for {Quadrotor} {Control}, Dec. 2023.
\newblock URL \url{http://arxiv.org/abs/2310.09053}.
\newblock arXiv:2310.09053 [cs].

\bibitem[Liao et~al.(2024)Liao, Zhang, Huang, Huang, Li, and Sreenath]{liao_berkeley_2024}
Q.~Liao, B.~Zhang, X.~Huang, X.~Huang, Z.~Li, and K.~Sreenath.
\newblock Berkeley {Humanoid}: {A} {Research} {Platform} for {Learning}-based {Control}, July 2024.
\newblock URL \url{http://arxiv.org/abs/2407.21781}.
\newblock arXiv:2407.21781 [cs].

\bibitem[Samak et~al.(2023)Samak, Samak, Kandhasamy, Krovi, and Xie]{samak_autodrive_2023}
T.~V. Samak, C.~V. Samak, S.~Kandhasamy, V.~Krovi, and M.~Xie.
\newblock {AutoDRIVE}: {A} {Comprehensive}, {Flexible} and {Integrated} {Digital} {Twin} {Ecosystem} for {Enhancing} {Autonomous} {Driving} {Research} and {Education}.
\newblock \emph{Robotics}, 12\penalty0 (3):\penalty0 77, May 2023.
\newblock ISSN 2218-6581.
\newblock \doi{10.3390/robotics12030077}.
\newblock URL \url{http://arxiv.org/abs/2212.05241}.
\newblock arXiv:2212.05241 [cs].

\bibitem[Williams et~al.(2017)Williams, Wagener, Goldfain, Drews, Rehg, Boots, and Theodorou]{williams_information_2017}
G.~Williams, N.~Wagener, B.~Goldfain, P.~Drews, J.~M. Rehg, B.~Boots, and E.~A. Theodorou.
\newblock Information theoretic {MPC} for model-based reinforcement learning.
\newblock In \emph{2017 {IEEE} {International} {Conference} on {Robotics} and {Automation} ({ICRA})}, pages 1714--1721, May 2017.
\newblock \doi{10.1109/ICRA.2017.7989202}.
\newblock URL \url{https://ieeexplore.ieee.org/document/7989202/?arnumber=7989202}.

\bibitem[Han et~al.(2024{\natexlab{a}})Han, Liu, Li, Spitzer, Shi, and Boots]{han_model_2024}
T.~Han, A.~Liu, A.~Li, A.~Spitzer, G.~Shi, and B.~Boots.
\newblock Model {Predictive} {Control} for {Aggressive} {Driving} {Over} {Uneven} {Terrain}, June 2024{\natexlab{a}}.
\newblock URL \url{http://arxiv.org/abs/2311.12284}.
\newblock arXiv:2311.12284 [cs].

\bibitem[Han et~al.(2024{\natexlab{b}})Han, Talia, Panicker, Shah, Jawale, and Boots]{han_dynamics_2024}
T.~Han, S.~Talia, R.~Panicker, P.~Shah, N.~Jawale, and B.~Boots.
\newblock Dynamics {Models} in the {Aggressive} {Off}-{Road} {Driving} {Regime}, May 2024{\natexlab{b}}.
\newblock URL \url{http://arxiv.org/abs/2405.16487}.
\newblock arXiv:2405.16487 [cs].

\bibitem[Datar et~al.(2024)Datar, Pan, Nazeri, and Xiao]{datar_toward_2024}
A.~Datar, C.~Pan, M.~Nazeri, and X.~Xiao.
\newblock Toward {Wheeled} {Mobility} on {Vertically} {Challenging} {Terrain}: {Platforms}, {Datasets}, and {Algorithms}.
\newblock In \emph{2024 {IEEE} {International} {Conference} on {Robotics} and {Automation} ({ICRA})}, pages 16322--16329, May 2024.
\newblock \doi{10.1109/ICRA57147.2024.10610079}.
\newblock URL \url{https://ieeexplore.ieee.org/document/10610079}.

\bibitem[Horváth et~al.(2024)Horváth, Ignéczi, Markó, Krecht, and Unger]{horvath_teaching_2024}
E.~Horváth, G.~Ignéczi, N.~Markó, R.~Krecht, and M.~Unger.
\newblock Teaching {Aspects} of {ROS} 2 and {Autonomous} {Vehicles}.
\newblock \emph{Engineering Proceedings}, 79\penalty0 (1):\penalty0 49, 2024.
\newblock ISSN 2673-4591.
\newblock \doi{10.3390/engproc2024079049}.
\newblock URL \url{https://www.mdpi.com/2673-4591/79/1/49}.
\newblock Number: 1 Publisher: Multidisciplinary Digital Publishing Institute.

\bibitem[Evans et~al.(2024)Evans, Trumpp, Caccamo, Jahncke, Betz, Jordaan, and Engelbrecht]{evans_unifying_2024}
B.~D. Evans, R.~Trumpp, M.~Caccamo, F.~Jahncke, J.~Betz, H.~W. Jordaan, and H.~A. Engelbrecht.
\newblock Unifying {F1TENTH} {Autonomous} {Racing}: {Survey}, {Methods} and {Benchmarks}, Apr. 2024.
\newblock URL \url{http://arxiv.org/abs/2402.18558}.
\newblock arXiv:2402.18558 [cs].

\bibitem[Dosovitskiy et~al.(2017)Dosovitskiy, Ros, Codevilla, Lopez, and Koltun]{dosovitskiy_carla_2017}
A.~Dosovitskiy, G.~Ros, F.~Codevilla, A.~Lopez, and V.~Koltun.
\newblock {CARLA}: {An} {Open} {Urban} {Driving} {Simulator}.
\newblock In \emph{Proceedings of the 1st {Annual} {Conference} on {Robot} {Learning}}, pages 1--16. PMLR, Oct. 2017.
\newblock URL \url{https://proceedings.mlr.press/v78/dosovitskiy17a.html}.
\newblock ISSN: 2640-3498.

\bibitem[Brunnbauer et~al.(2022)Brunnbauer, Berducci, Brandstätter, Lechner, Hasani, Rus, and Grosu]{brunnbauer_latent_2022}
A.~Brunnbauer, L.~Berducci, A.~Brandstätter, M.~Lechner, R.~Hasani, D.~Rus, and R.~Grosu.
\newblock Latent {Imagination} {Facilitates} {Zero}-{Shot} {Transfer} in {Autonomous} {Racing}, Feb. 2022.
\newblock URL \url{http://arxiv.org/abs/2103.04909}.
\newblock arXiv:2103.04909 [cs].

\bibitem[Hamilton et~al.(2022)Hamilton, Musau, Lopez, and Johnson]{hamilton_zero-shot_2022}
N.~Hamilton, P.~Musau, D.~M. Lopez, and T.~T. Johnson.
\newblock Zero-{Shot} {Policy} {Transfer} in {Autonomous} {Racing}: {Reinforcement} {Learning} vs {Imitation} {Learning}.
\newblock In \emph{2022 {IEEE} {International} {Conference} on {Assured} {Autonomy} ({ICAA})}, pages 11--20, Mar. 2022.
\newblock \doi{10.1109/ICAA52185.2022.00011}.
\newblock URL \url{https://ieeexplore.ieee.org/document/9763640/?arnumber=9763640}.

\bibitem[Cai et~al.(2025)Cai, Queeney, Xu, Datar, Pan, Miller, Flather, Osteen, Roy, Xiao, and How]{cai_pietra_2025}
X.~Cai, J.~Queeney, T.~Xu, A.~Datar, C.~Pan, M.~Miller, A.~Flather, P.~R. Osteen, N.~Roy, X.~Xiao, and J.~P. How.
\newblock {PIETRA}: {Physics}-{Informed} {Evidential} {Learning} for {Traversing} {Out}-of-{Distribution} {Terrain}.
\newblock \emph{IEEE Robotics and Automation Letters}, pages 1--8, 2025.
\newblock ISSN 2377-3766.
\newblock \doi{10.1109/LRA.2025.3527285}.
\newblock URL \url{https://ieeexplore.ieee.org/document/10833878/?arnumber=10833878}.
\newblock Conference Name: IEEE Robotics and Automation Letters.

\bibitem[Deitke et~al.(2020)Deitke, Han, Herrasti, Kembhavi, Kolve, Mottaghi, Salvador, Schwenk, VanderBilt, Wallingford, Weihs, Yatskar, and Farhadi]{deitke_robothor_2020}
M.~Deitke, W.~Han, A.~Herrasti, A.~Kembhavi, E.~Kolve, R.~Mottaghi, J.~Salvador, D.~Schwenk, E.~VanderBilt, M.~Wallingford, L.~Weihs, M.~Yatskar, and A.~Farhadi.
\newblock {RoboTHOR}: {An} {Open} {Simulation}-to-{Real} {Embodied} {AI} {Platform}.
\newblock In \emph{2020 {IEEE}/{CVF} {Conference} on {Computer} {Vision} and {Pattern} {Recognition} ({CVPR})}, pages 3161--3171, Seattle, WA, USA, June 2020. IEEE.
\newblock ISBN 978-1-72817-168-5.
\newblock \doi{10.1109/CVPR42600.2020.00323}.
\newblock URL \url{https://ieeexplore.ieee.org/document/9157346/}.

\bibitem[Balaji et~al.(2020)Balaji, Mallya, Genc, Gupta, Dirac, Khare, Roy, Sun, Tao, Townsend, Calleja, Muralidhara, and Karuppasamy]{balaji_deepracer_2020}
B.~Balaji, S.~Mallya, S.~Genc, S.~Gupta, L.~Dirac, V.~Khare, G.~Roy, T.~Sun, Y.~Tao, B.~Townsend, E.~Calleja, S.~Muralidhara, and D.~Karuppasamy.
\newblock {DeepRacer}: {Autonomous} {Racing} {Platform} for {Experimentation} with {Sim2Real} {Reinforcement} {Learning}.
\newblock In \emph{2020 {IEEE} {International} {Conference} on {Robotics} and {Automation} ({ICRA})}, pages 2746--2754, May 2020.
\newblock \doi{10.1109/ICRA40945.2020.9197465}.
\newblock URL \url{https://ieeexplore.ieee.org/abstract/document/9197465?casa_token=P1_UodV4wbEAAAAA:6EV0TBoDKvtpQailXF-0QMLLebuYMajCFlRjbSzrKQB__Q2khTqumEjWZKAM4C3Xrg7WgicvRg}.
\newblock ISSN: 2577-087X.

\bibitem[Salimpour et~al.(2025)Salimpour, Peña-Queralta, Paez-Granados, Heikkonen, and Westerlund]{salimpour_sim--real_2025}
S.~Salimpour, J.~Peña-Queralta, D.~Paez-Granados, J.~Heikkonen, and T.~Westerlund.
\newblock Sim-to-{Real} {Transfer} for {Mobile} {Robots} with {Reinforcement} {Learning}: from {NVIDIA} {Isaac} {Sim} to {Gazebo} and {Real} {ROS} 2 {Robots}, Jan. 2025.
\newblock URL \url{http://arxiv.org/abs/2501.02902}.
\newblock arXiv:2501.02902 [cs].

\bibitem[Samak et~al.(2023)Samak, Samak, and Krovi]{samak_towards_2023}
C.~Samak, T.~Samak, and V.~Krovi.
\newblock Towards {Sim2Real} {Transfer} of {Autonomy} {Algorithms} using {AutoDRIVE} {Ecosystem}.
\newblock \emph{IFAC-PapersOnLine}, 56\penalty0 (3):\penalty0 277--282, Jan. 2023.
\newblock ISSN 2405-8963.
\newblock \doi{10.1016/j.ifacol.2023.12.037}.
\newblock URL \url{https://www.sciencedirect.com/science/article/pii/S2405896323023704}.

\bibitem[Xiao et~al.(2024)Xiao, Xue, Tao, Kalaria, Dolan, and Shi]{xiao_anycar_2024}
W.~Xiao, H.~Xue, T.~Tao, D.~Kalaria, J.~M. Dolan, and G.~Shi.
\newblock {AnyCar} to {Anywhere}: {Learning} {Universal} {Dynamics} {Model} for {Agile} and {Adaptive} {Mobility}, Sept. 2024.
\newblock URL \url{http://arxiv.org/abs/2409.15783}.
\newblock arXiv:2409.15783 [cs].

\bibitem[Cutler and How(2016)]{cutler_autonomous_2016}
M.~Cutler and J.~P. How.
\newblock Autonomous drifting using simulation-aided reinforcement learning.
\newblock In \emph{2016 {IEEE} {International} {Conference} on {Robotics} and {Automation} ({ICRA})}, pages 5442--5448, Stockholm, Sweden, May 2016. IEEE.
\newblock ISBN 978-1-4673-8026-3.
\newblock \doi{10.1109/ICRA.2016.7487756}.
\newblock URL \url{http://ieeexplore.ieee.org/document/7487756/}.

\bibitem[Evans et~al.(2023)Evans, Jordaan, and Engelbrecht]{evans_comparing_2023}
B.~D. Evans, H.~W. Jordaan, and H.~A. Engelbrecht.
\newblock Comparing deep reinforcement learning architectures for autonomous racing.
\newblock \emph{Machine Learning with Applications}, 14:\penalty0 100496, Dec. 2023.
\newblock ISSN 26668270.
\newblock \doi{10.1016/j.mlwa.2023.100496}.
\newblock URL \url{https://linkinghub.elsevier.com/retrieve/pii/S266682702300049X}.

\bibitem[Gonzales et~al.(2016)Gonzales, Zhang, Li, and Borrelli]{gonzales_autonomous_2016}
J.~Gonzales, F.~Zhang, K.~Li, and F.~Borrelli.
\newblock Autonomous drifting with onboard sensors.
\newblock In \emph{Advanced {Vehicle} {Control}}. CRC Press, 2016.
\newblock ISBN 978-1-315-26528-5.
\newblock Num Pages: 6.

\bibitem[Cai et~al.(2020)Cai, Mei, Tai, Sun, and Liu]{cai_high-speed_2020}
P.~Cai, X.~Mei, L.~Tai, Y.~Sun, and M.~Liu.
\newblock High-{Speed} {Autonomous} {Drifting} {With} {Deep} {Reinforcement} {Learning}.
\newblock \emph{IEEE Robotics and Automation Letters}, 5\penalty0 (2):\penalty0 1247--1254, Apr. 2020.
\newblock ISSN 2377-3766.
\newblock \doi{10.1109/LRA.2020.2967299}.
\newblock URL \url{https://ieeexplore.ieee.org/document/8961997/?arnumber=8961997}.
\newblock Conference Name: IEEE Robotics and Automation Letters.

\bibitem[Djeumou et~al.(2024)Djeumou, Thompson, Suminaka, and Subosits]{djeumou_reference-free_2024}
F.~Djeumou, M.~Thompson, M.~Suminaka, and J.~Subosits.
\newblock Reference-{Free} {Formula} {Drift} with {Reinforcement} {Learning}: {From} {Driving} {Data} to {Tire} {Energy}-{Inspired}, {Real}-{World} {Policies}, Oct. 2024.
\newblock URL \url{http://arxiv.org/abs/2410.20990}.
\newblock arXiv:2410.20990 [cs].

\bibitem[Chen et~al.(2023)Chen, Zhao, Gao, and Hua]{chen_dynamic_2023}
G.~Chen, X.~Zhao, Z.~Gao, and M.~Hua.
\newblock Dynamic {Drifting} {Control} for {General} {Path} {Tracking} of {Autonomous} {Vehicles}.
\newblock \emph{IEEE Transactions on Intelligent Vehicles}, 8\penalty0 (3):\penalty0 2527--2537, Mar. 2023.
\newblock ISSN 2379-8904.
\newblock \doi{10.1109/TIV.2023.3235007}.
\newblock URL \url{https://ieeexplore.ieee.org/document/10011537/?arnumber=10011537}.
\newblock Conference Name: IEEE Transactions on Intelligent Vehicles.

\bibitem[Frey et~al.(2024)Frey, Patel, Atha, Nubert, Fan, Agha, Padgett, Spieler, Hutter, and Khattak]{frey_roadrunner_2024}
J.~Frey, M.~Patel, D.~Atha, J.~Nubert, D.~Fan, A.~Agha, C.~Padgett, P.~Spieler, M.~Hutter, and S.~Khattak.
\newblock {RoadRunner} -- {Learning} {Traversability} {Estimation} for {Autonomous} {Off}-road {Driving}, Aug. 2024.
\newblock URL \url{http://arxiv.org/abs/2402.19341}.
\newblock arXiv:2402.19341 [cs].

\bibitem[Gibson et~al.(2023)Gibson, Vlahov, Fan, Spieler, Pastor, Agha-mohammadi, and Theodorou]{gibson_multi-step_2023}
J.~Gibson, B.~Vlahov, D.~Fan, P.~Spieler, D.~Pastor, A.-a. Agha-mohammadi, and E.~A. Theodorou.
\newblock A {Multi}-step {Dynamics} {Modeling} {Framework} {For} {Autonomous} {Driving} {In} {Multiple} {Environments}, May 2023.
\newblock URL \url{http://arxiv.org/abs/2305.02241}.
\newblock arXiv:2305.02241 [cs].

\bibitem[Stachowicz et~al.(2023)Stachowicz, Shah, Bhorkar, Kostrikov, and Levine]{stachowicz_fastrlap_2023}
K.~Stachowicz, D.~Shah, A.~Bhorkar, I.~Kostrikov, and S.~Levine.
\newblock {FastRLAP}: {A} {System} for {Learning} {High}-{Speed} {Driving} via {Deep} {RL} and {Autonomous} {Practicing}.
\newblock In \emph{Proceedings of {The} 7th {Conference} on {Robot} {Learning}}, pages 3100--3111. PMLR, Dec. 2023.
\newblock URL \url{https://proceedings.mlr.press/v229/stachowicz23a.html}.
\newblock ISSN: 2640-3498.

\bibitem[Xu et~al.(2024)Xu, Pan, and Xiao]{xu_reinforcement_2024}
T.~Xu, C.~Pan, and X.~Xiao.
\newblock Reinforcement {Learning} for {Wheeled} {Mobility} on {Vertically} {Challenging} {Terrain}.
\newblock In \emph{2024 {IEEE} {International} {Symposium} on {Safety} {Security} {Rescue} {Robotics} ({SSRR})}, pages 125--130, Nov. 2024.
\newblock \doi{10.1109/SSRR62954.2024.10770034}.
\newblock URL \url{https://ieeexplore.ieee.org/document/10770034/?arnumber=10770034}.
\newblock ISSN: 2475-8426.

\bibitem[Kang et~al.(2019)Kang, Belkhale, Kahn, Abbeel, and Levine]{kang_generalization_2019}
K.~Kang, S.~Belkhale, G.~Kahn, P.~Abbeel, and S.~Levine.
\newblock Generalization through {Simulation}: {Integrating} {Simulated} and {Real} {Data} into {Deep} {Reinforcement} {Learning} for {Vision}-{Based} {Autonomous} {Flight}.
\newblock In \emph{2019 {International} {Conference} on {Robotics} and {Automation} ({ICRA})}, pages 6008--6014, May 2019.
\newblock \doi{10.1109/ICRA.2019.8793735}.
\newblock URL \url{https://ieeexplore.ieee.org/document/8793735/?arnumber=8793735}.
\newblock ISSN: 2577-087X.

\bibitem[Yuan et~al.(2024)Yuan, Wei, Cheng, Zhang, Chen, and Xu]{yuan_learning_2024}
Z.~Yuan, T.~Wei, S.~Cheng, G.~Zhang, Y.~Chen, and H.~Xu.
\newblock Learning to {Manipulate} {Anywhere}: {A} {Visual} {Generalizable} {Framework} {For} {Reinforcement} {Learning}, Oct. 2024.
\newblock URL \url{http://arxiv.org/abs/2407.15815}.
\newblock arXiv:2407.15815 [cs].

\bibitem[Technologies()]{technologies_1x_nodate}
X.~Technologies.
\newblock {1X} {Technologies} {\textbar} {Safe}, {Intelligent} {Humanoids}.
\newblock URL \url{https://www.1x.tech/}.

\bibitem[noa({\natexlab{a}})]{noauthor_crossing_nodate}
Crossing the {Sim2Real} {Gap} {With} {NVIDIA} {Isaac} {Lab}, {\natexlab{a}}.
\newblock URL \url{https://agilityrobotics.com/content/crossing-sim2real-gap-with-isaaclab}.

\bibitem[noa({\natexlab{b}})]{noauthor_get_nodate}
Get {Started} with {Reinforcement} {Learning} for {Spot}, {\natexlab{b}}.
\newblock URL \url{https://support.bostondynamics.com/s/article/Get-Started-with-Reinforcement-Learning-for-Spot-49966}.

\bibitem[noa({\natexlab{c}})]{noauthor_nvidia_nodate}
{NVIDIA} {Isaac} {Lab} {Blog}, {\natexlab{c}}.
\newblock URL \url{https://fieldai.com/news/field-ai-nvidia-partnership}.

\bibitem[noa({\natexlab{d}})]{noauthor_menteebot_nodate}
{MenteeBot}, {\natexlab{d}}.
\newblock URL \url{https://menteebot.com/blog/#shopping-companion-2024}.

\bibitem[Ma et~al.(2024)Ma, Liang, Wang, Huang, Bastani, Jayaraman, Zhu, Fan, and Anandkumar]{ma_eureka_2024}
Y.~J. Ma, W.~Liang, G.~Wang, D.-A. Huang, O.~Bastani, D.~Jayaraman, Y.~Zhu, L.~Fan, and A.~Anandkumar.
\newblock Eureka: {Human}-{Level} {Reward} {Design} via {Coding} {Large} {Language} {Models}, Apr. 2024.
\newblock URL \url{http://arxiv.org/abs/2310.12931}.
\newblock arXiv:2310.12931 [cs].

\bibitem[Tao et~al.(2024)Tao, Xiang, Shukla, Qin, Hinrichsen, Yuan, Bao, Lin, Liu, Chan, Gao, Li, Mu, Xiao, Gurha, Huang, Calandra, Chen, Luo, and Su]{tao_maniskill3_2024}
S.~Tao, F.~Xiang, A.~Shukla, Y.~Qin, X.~Hinrichsen, X.~Yuan, C.~Bao, X.~Lin, Y.~Liu, T.-k. Chan, Y.~Gao, X.~Li, T.~Mu, N.~Xiao, A.~Gurha, Z.~Huang, R.~Calandra, R.~Chen, S.~Luo, and H.~Su.
\newblock {ManiSkill3}: {GPU} {Parallelized} {Robotics} {Simulation} and {Rendering} for {Generalizable} {Embodied} {AI}, Oct. 2024.
\newblock URL \url{http://arxiv.org/abs/2410.00425}.
\newblock arXiv:2410.00425 [cs].

\bibitem[Yang et~al.(2024)Yang, Shi, Lin, Meng, Scalise, Castro, Yu, Zhang, Zhao, Tan, and Boots]{yang_agile_2024}
Y.~Yang, G.~Shi, C.~Lin, X.~Meng, R.~Scalise, M.~G. Castro, W.~Yu, T.~Zhang, D.~Zhao, J.~Tan, and B.~Boots.
\newblock Agile {Continuous} {Jumping} in {Discontinuous} {Terrains}, Sept. 2024.
\newblock URL \url{http://arxiv.org/abs/2409.10923}.
\newblock arXiv:2409.10923 [cs].

\bibitem[Han et~al.(2025)Han, Bao, Mehta, Guo, Vishwakarma, Kang, Jung, Scalise, Zhou, Xu, and Boots]{han_model_2025}
T.~Han, Y.~Bao, B.~Mehta, G.~Guo, A.~Vishwakarma, E.~Kang, S.~Jung, R.~Scalise, J.~Zhou, B.~Xu, and B.~Boots.
\newblock Model {Predictive} {Adversarial} {Imitation} {Learning} for {Planning} from {Observation}, July 2025.
\newblock URL \url{http://arxiv.org/abs/2507.21533}.
\newblock arXiv:2507.21533 [cs].

\bibitem[Schulman et~al.(2017)Schulman, Wolski, Dhariwal, Radford, and Klimov]{schulman_proximal_2017}
J.~Schulman, F.~Wolski, P.~Dhariwal, A.~Radford, and O.~Klimov.
\newblock Proximal {Policy} {Optimization} {Algorithms}, Aug. 2017.
\newblock URL \url{http://arxiv.org/abs/1707.06347}.
\newblock arXiv:1707.06347.

\bibitem[Lee et~al.(2020)Lee, Hwangbo, Wellhausen, Koltun, and Hutter]{lee_learning_2020}
J.~Lee, J.~Hwangbo, L.~Wellhausen, V.~Koltun, and M.~Hutter.
\newblock Learning quadrupedal locomotion over challenging terrain.
\newblock \emph{Science Robotics}, 5\penalty0 (47):\penalty0 eabc5986, Oct. 2020.
\newblock \doi{10.1126/scirobotics.abc5986}.
\newblock URL \url{https://www.science.org/doi/10.1126/scirobotics.abc5986}.
\newblock Publisher: American Association for the Advancement of Science.

\bibitem[Acosta et~al.(2017)Acosta, Kanarachos, and Fitzpatrick]{acosta_hybrid_2017}
M.~Acosta, S.~Kanarachos, and M.~E. Fitzpatrick.
\newblock A {Hybrid} {Hierarchical} {Rally} {Driver} {Model} for {Autonomous} {Vehicle} {Agile} {Maneuvering} on {Loose} {Surfaces}:.
\newblock In \emph{Proceedings of the 14th {International} {Conference} on {Informatics} in {Control}, {Automation} and {Robotics}}, pages 216--225, Madrid, Spain, 2017. SCITEPRESS - Science and Technology Publications.
\newblock ISBN 978-989-758-263-9 978-989-758-264-6.
\newblock \doi{10.5220/0006393002160225}.
\newblock URL \url{http://www.scitepress.org/DigitalLibrary/Link.aspx?doi=10.5220/0006393002160225}.

\bibitem[Jung et~al.(2024)Jung, Lee, Meng, Boots, and Lambert]{jung_v-strong_2024}
S.~Jung, J.~Lee, X.~Meng, B.~Boots, and A.~Lambert.
\newblock V-{STRONG}: {Visual} {Self}-{Supervised} {Traversability} {Learning} for {Off}-road {Navigation}.
\newblock In \emph{2024 {IEEE} {International} {Conference} on {Robotics} and {Automation} ({ICRA})}, pages 1766--1773, May 2024.
\newblock \doi{10.1109/ICRA57147.2024.10611227}.
\newblock URL \url{https://ieeexplore.ieee.org/abstract/document/10611227}.

\bibitem[Yasuda et~al.(2021)Yasuda, Martins, and Cappabianco]{yasuda_autonomous_2021}
Y.~D.~V. Yasuda, L.~E.~G. Martins, and F.~A.~M. Cappabianco.
\newblock Autonomous {Visual} {Navigation} for {Mobile} {Robots}: {A} {Systematic} {Literature} {Review}.
\newblock \emph{ACM Computing Surveys}, 53\penalty0 (1):\penalty0 1--34, Jan. 2021.
\newblock ISSN 0360-0300, 1557-7341.
\newblock \doi{10.1145/3368961}.
\newblock URL \url{https://dl.acm.org/doi/10.1145/3368961}.

\bibitem[Choi et~al.(2021)Choi, Jung, Yun, Kim, Kim, and Choo]{choi_robustnet_2021}
S.~Choi, S.~Jung, H.~Yun, J.~T. Kim, S.~Kim, and J.~Choo.
\newblock {RobustNet}: {Improving} {Domain} {Generalization} in {Urban}-{Scene} {Segmentation} via {Instance} {Selective} {Whitening}.
\newblock pages 11580--11590, 2021.
\newblock URL \url{https://openaccess.thecvf.com/content/CVPR2021/html/Choi_RobustNet_Improving_Domain_Generalization_in_Urban-Scene_Segmentation_via_Instance_Selective_CVPR_2021_paper.html}.

\bibitem[Bousmalis et~al.(2018)Bousmalis, Irpan, Wohlhart, Bai, Kelcey, Kalakrishnan, Downs, Ibarz, Pastor, Konolige, Levine, and Vanhoucke]{bousmalis_using_2018}
K.~Bousmalis, A.~Irpan, P.~Wohlhart, Y.~Bai, M.~Kelcey, M.~Kalakrishnan, L.~Downs, J.~Ibarz, P.~Pastor, K.~Konolige, S.~Levine, and V.~Vanhoucke.
\newblock Using {Simulation} and {Domain} {Adaptation} to {Improve} {Efficiency} of {Deep} {Robotic} {Grasping}.
\newblock In \emph{2018 {IEEE} {International} {Conference} on {Robotics} and {Automation} ({ICRA})}, pages 4243--4250, May 2018.
\newblock \doi{10.1109/ICRA.2018.8460875}.
\newblock URL \url{https://ieeexplore.ieee.org/document/8460875/?arnumber=8460875}.
\newblock ISSN: 2577-087X.

\end{thebibliography}
